\documentclass[10pt,journal,compsoc]{IEEEtran}

\ifCLASSOPTIONcompsoc
  \usepackage[nocompress]{cite}
\else
  \usepackage{cite}
\fi

\usepackage[utf8]{inputenc} 
\usepackage[T1]{fontenc}    
\usepackage{hyperref}       
\usepackage{url}            
\usepackage{booktabs}       
\usepackage{amsfonts}       
\usepackage{nicefrac}       
\usepackage{microtype}      
\usepackage{lipsum}
\usepackage{graphicx}
\usepackage{amsmath}
\usepackage{amssymb}
\usepackage{rotating}
\usepackage{optidef}
\usepackage{multicol}
\usepackage{tabularx}

\DeclareMathOperator*{\argmax}{arg\,max}
\DeclareMathOperator*{\argmin}{arg\,min}
\usepackage{algorithm}
\usepackage[noend]{algpseudocode}

\usepackage{tikz}
\usetikzlibrary{arrows,shapes,positioning,shadows,trees}

\newcommand{\mynorm}[1]{\left \| #1 \right \|}

\IEEEoverridecommandlockouts

\usepackage{enumitem}
\setlist{leftmargin=*}

\tikzset{
  basic/.style  = {draw, text width=2cm, drop shadow, font=\sffamily, rectangle},
  root/.style   = {basic, rounded corners=2pt, thin, align=center,
                   fill=gray!30},
  level 2/.style = {basic, rounded corners=6pt, thin,align=center, fill=gray!20,
                   text width=8em},
  level 3/.style = {basic, thin, align=left, fill=gray!10, text width=6.5em}
}

\title{Opportunities and Challenges in Deep Learning Adversarial Robustness: A Survey}

\author{Samuel~Henrique~Silva,~\IEEEmembership{Member,~IEEE,}
        and~Peyman~Najafirad,~\IEEEmembership{Senior,~IEEE}
\IEEEcompsocitemizethanks{\IEEEcompsocthanksitem S.H. Silva and P. Najafirad are members of the Secure AI \& Autonomy Laboratory, with the Department
of Electrical and Computer Engineering, University of Texas at San Antonio, San Antonio,
TX, 78249.\protect\\
\IEEEcompsocthanksitem P. Najafirad (corresponding author) is also with Department of Information Systems and Cyber Security, University of Texas at San Antonio, San Antonio, TX, 78249.\protect\\
E-mail: peyman.najafirad@utsa.edu}
\thanks{The authors gratefully acknowledge the use of the services of Jetstream cloud, funded by National Science Foundation, United States award 1445604.}
\thanks{This work has been submitted to the IEEE for possible publication. Copyright may be transferred without notice, after which this version may no longer be accessible.}}

\begin{document}

\IEEEtitleabstractindextext{%
\begin{abstract}
As we seek to deploy machine learning models beyond virtual and controlled domains, it is critical to analyze not only the accuracy or the fact that it works most of the time, but if such a model is truly robust and reliable. This paper studies strategies to implement adversary robustly trained algorithms towards guaranteeing safety in machine learning algorithms. We provide a taxonomy to classify adversarial attacks and defenses, formulate the Robust Optimization problem in a min-max setting, and divide it into 3 subcategories, namely: Adversarial (re)Training, Regularization Approach, and Certified Defenses. We survey the most recent and important results in adversarial example generation, defense mechanisms with adversarial (re)Training as their main defense against perturbations. We also survey mothods that add regularization terms which change the behavior of the gradient, making it harder for attackers to achieve their objective. Alternatively, we've surveyed methods which formally derive certificates of robustness by exactly solving the optimization problem or by approximations using upper or lower bounds. In addition we discuss the challenges faced by most of the recent algorithms presenting future research perspectives.

\end{abstract}

}
\maketitle


\section{Introduction}
\label{sec:intro}
Deep Learning (DL) (\cite{lecun2015deep}) models are changing the way we solve problems that have required many attempts from the most diverse fields of science. DL is an improvement over Artificial Intelligence (AI) Neural Networks (NN), in which more layers are stacked to grant a bigger level of abstraction and better reasoning over the data when compared to other Machine Learning (ML) algorithms (\cite{greenspan2016guest}). Since the raise of DL, supported in many cases by cloud environments \cite{stewart2015jetstream, towns2014xsede, das2018distributed}, the base architecture and its variations have been applied in many scientific breakthroughs in the most diverse fields of knowledge, e.g. in predicting AMD disease progression (\cite{das2019distributed}),  predicting DNA enhancers for gene expression programmes (\cite{yang2017biren}), elections and demographic analysis based on satellite images (\cite{gebru2017using}), filtering data for gravitational-wave signals (\cite{gabbard2018matching}). DL approach has also become one of the most used approaches for natural language processing (\cite{ebadi2019implicit}) and speech recognition (\cite{amodei2016deep}).

One of the most popular variations of DL architecture, Convolutional Neural Networks (CNN) have significantly boosted the performance of DL algorithms in computer vision (CV) applications (\cite{voulodimos2018deep}), bringing it to several areas of CV such as, object detection (\cite{redmon2016you, dai2016r}, action recognition \cite{varol2017long, 9043480}, pose estimation \cite{cao2017realtime, newell2016stacked}, image segmentation \cite{badrinarayanan2017segnet, chen2017deeplab}, and motion tracking \cite{held2016learning}. Starting with ImageNet \cite{krizhevsky2012imagenet}, proposed in 2012, the field of CNN's have seen great improvement with super-human performance in specific tasks, providing solutions even to medical problems \cite{litjens2017survey}.

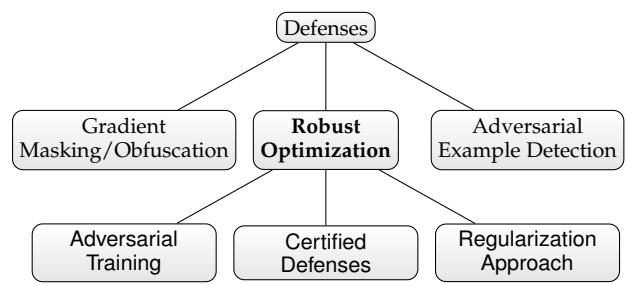
\begin{figure}
    \centering
    \begin{tikzpicture}[sibling distance=8em,
  every node/.style = {shape=rectangle, rounded corners,
    draw, align=center,
    top color=white, bottom color=gray!20}]]
    \footnotesize
  \node {Defenses}
    child { node {Gradient\\Masking/Obfuscation} }
    child { node {\textbf{Robust}\\\textbf{Optimization}}
        child { node {Adversarial\\Training}}
        child { node {Certified\\Defenses}}
        child { node {Regularization\\Approach}}}
    child { node {Adversarial\\Example Detection} };
\end{tikzpicture}
    \caption{Defenses against adversarial attacks are divided in 3 categories: Gradient Masking/Obfuscation, Robust Optimization, and Adversarial Example Detection. The focus of this survey is Robust Optimization which we subdivide in: Adversarial Training, Certified Defenses, and Regularization Approach.}
    \label{fig:defenses_adv_attack_generic}
\end{figure}

Fueled by the fact that new frameworks, libraries, and hardware resources are being improved and made available to the public and scientific community \cite{abadi2016tensorflow, paszke2019pytorch, keahey2019chameleon}, Deep Neural networks (DNN) are being improved constantly and achieving new performance breakthroughs \cite{he2016deep, szegedy2016rethinking, szegedy2017inception}. With the current maturity of DNN algorithms, its being applied in solving safety and security-critical problems  \cite{akhtar2018threat}, such as self-driving cars \cite{bojarski2016end, silva2017multi}, multi-agent aerial vehicle systems with face identification \cite{silva2019cooperative}, robotics \cite{gu2017deep, silva2020temporal}, social engineering detection \cite{lansley2019seen}, network anomaly detection \cite{kwon2017survey}, deep packet inspection in networks \cite{de2019implementation}. DNN applications are already part of our day-to-day life (personal assistants \cite{kepuska2018next}, product recommendation \cite{cheng2016wide}, biometric identification \cite{de2018driverless}) and tend to occupy a bigger space as time passes.

As seen in many publications, DNN has been shown to have human-level accuracy even for significantly complex tasks such as playing games with no prior rule known, except the current frames \cite{mnih2015human}. In contrast to the aforementioned accuracy of DNN models, its been shown in earlier publications \cite{szegedy2013intriguing, nguyen2015deep, biggio2014security}, that DNN models are susceptible to small input perturbations, in most cases imperceptible to the human eye. The results from this publications have shown the facility with which small additive targeted noise to the input image, makes models to misclassify objects which before could be identified with 99.99\% confidence. More alarming is the fact that such models report high confidence in the predictions. Such perturbations, which can fool a trained model, are known as adversarial attacks. With such alarming consequences, the study of adversarial attacks and robustness against them became a great deal of research in recent years.

A considerably large number of research papers is now available concerning methods to identify adversarial attacks and defend from incursions against the model \cite{chakraborty2018adversarial, chacon2019deep}. One way of solving this issue is adding better intuition on the models, through explainability \cite{das2020opportunities}, but such models do not target the direct improvement of the model. On the other hand, several approaches have been published to generate models which are robust against adversarial attacks \cite{carlini2017towards}, the target of the researchers is to introduce in their models' layers of robustness such that the models are not fooled by out of distribution examples, known or unknown attacks, targeted or untargeted attacks. Guaranteeing the accuracy of such models while safety is taken into consideration, is of utmost importance for system architects, mainly making them robust to the presence of adversarial attacks, noise, model misspecification, and uncertainty. This survey aims to bring together the recent advances in robustness for DNN's, pointing the main research directions being followed recently to generate robust DNN models. We bring to light both applied and theoretical recent developments.

Inspired by \cite{yuan2019adversarial}, we analyze the robustness of DNN's under the perspective of how adversarial examples can be generated, and how defenses can be formulated against such algorithms. In general words, we define the robustness against adversarial attacks problem as a dual optimization problem, in which the attackers try to maximize the loss while the defenses try to minimize the chance of a model being fooled by the attacker. In such formulation, current existing models based in non-linear activation functions, introduce non-linear inequality constraints in the optimization, which generates an inherent trade-off between exact solutions and scalability of the model. This trade-off comes in the form of either exact slow solutions through mixed-integer linear programming, or in approximations to the objective function which either, relies on the existing attack methods to provide a local heuristic estimation to the maximization function, or approximations of the bounds of the constraints or objective function to generate certification regions, in which no adversarial example exists. 

More specifically this paper presents the following contributions:

\begin{enumerate}
    \item We characterize defenses to adversarial attacks as a min-max optimization problem, investigating solutions involving heuristic approximation, exact solution, and upper/lower bound approximations to generate models robust to adversarial attacks.
    \item We investigate, analyze, and categorize the most recent and/or important approaches to generate adversarial examples, as they are the basis to generate strong defenses, through Adversarial (re)Training. 
    \item We investigate, analyze, and categorize the most recent and important approaches to generate defenses against adversarial attacks, providing a taxonomy, description of the methods, and the main results in such approaches.
\end{enumerate}

We organize this survey in the following manner. In \autoref{sec:Taxonomy_adv} we describe taxonomies for both adversarial example generation and defenses. We classify the adversarial models concerning the time of the attack, information available to the attacker, objective, and the algorithm computation method. Moreover, we classify the perturbation type used by the attackers. We divide the defense methods into three categories, namely gradient-masking/obfuscation, robust optimization, and adversarial example detection. We focus this research in Robust Optimization, and further sub-divide in 3 groups: Adversarial Training,  Certified Defenses, and Regularization Approach. In \autoref{sec:adversarial_attacks}, we describe several relevant adversarial attacks, and summarize them in \autoref{tab:adv_att_table}. In \autoref{sec:robust_optimization} we describe the most relevant results in Robust Optimization, and provide a tree that maps these publications to the 3 sub-groups of Robust Optimization. In \autoref{sec:ooportunities_challenges} we discuss current challenges and opportunities in robust defenses.

\section{Taxonomy of Adversarial Attacks and Defenses}
\label{sec:Taxonomy_adv}
We keep a consistent notation set along with the survey, and for easiness of reading we summarize in \autoref{tab:notations} the most used notations and symbols which we will use along with this survey. For papers requiring some specific terms, we define them in the section in which they are presented.
\begin{table}[!ht]
    
    \caption{Symbols and notations used in the mathematical definitions} 
    \centering 
    \begin{tabular}{|c| l|} %
    \hline  
    Symbol & Description \\ [0.5ex] 
    \hline 
    \hline
    $x$  & original (clean, unmodified) input data \\\hline
    $\hat{y}$ & model's prediction \\\hline
    $t$  & class label \\\hline
    $x'$ & adversarial example \\\hline
    $y'$ & target class of adversarial example \\\hline
    $f(.)$ & DL model \\\hline
    $\theta$ & parameters of the model \\\hline
    $\delta$ & perturbation generated by adv. algorithm \\\hline
    $\Delta$, $\epsilon$ & perturbation constraint \\\hline
    $\nabla$ & gradient function \\\hline
    $\left \| . \right \|_p$ & the $l_p$-norm \\\hline
    $\mathcal{L}$ & loss function (e.g., cross-entropy) \\\hline
    $\mathcal{D}$ & Training data distribution \\\hline
    $KL$-divergence & Kullback-Leibler divergence function\\\hline
    \end{tabular}
    \label{tab:notations}
\end{table}
\subsection{Attack Threat Model}

Several attempts have been made to categorize attacks on machine learning. We here distill the most important aspects which characterize adversarial examples generating models concerning their architecture. We focus on the aspects that are most relevant to the discussion of adversarial robustness. To that end, we classify the attacks concerning timing, information, goals, and attack frequency following the proposed in \cite{vorobeychik2018adversarial}:
\begin{itemize}[leftmargin=*]
    \item \textbf{Timing:} A first crucial feature for modeling the adversarial attacks is when it occurs. To that end we have two possibilities, \textit{evasion} and \textit{poisoning} attacks. Evasion attacks are the ones in the time of inference and assume the model has already been trained. Poisoning attacks in general targets the data, and the training phase of the model.
    
    \item \textbf{Information:} Another feature of the attack references the information to which the attacker has access. In the \textit{white box} context the adversary has full access to information concerning the model and the model itself, as opposed to \textit{black box} setting, in which very few or no information is available. White box attacks refer to those in which the adversary can unrestrictedly query the model for any information, such as weights, gradient, model hyper-parameters, prediction scores. Whereas in black-box attacks the adversary has limited or no information about these parameters, although may obtain some of the information indirectly, for example, through queries. Some also define \textit{grey-box attacks}, in which attackers might only know the feature representation and the type of model that was used but have no access to dataset or the model information.  A fourth setting is called \textit{restricted black-box}, or also known as \textit{no-box attack}. Under such an assumption, no information is available to the attacker, and the research is mainly focused on attack transferability. In wich the focus is to evaluate the possibility of transferring the attack performed in one DNN to the inaccessible objective model \cite{chen2017zoo}. In this work, we evaluate models in a binary setting, the adversary either has  comprehensive access to the DNN or black box having limited access through queries, which can also provide class scores.
    
    \item \textbf{Goals:} The attackers may have different reasons to target a specific algorithm. But mostly the attacker has either a specific goal, and needs the algorithm to output a specific output, case in which it is a \textit{targeted attack}, or just wants to reduce the reliability of the algorithm by forcing a mistake. In the latter, we have an untargeted attack.
    
    \item \textbf{Attack Frequency:} The attack on the victim's model can be either iterative or one-time. In the one-time, the optimization of the objective function of the attacker happens in a single step, whereas the iterative method takes several steps to generate the perturbation. 
    
\end{itemize}

\subsection{Attack Perturbation Type}

The size of the perturbation is in the core of the adversarial attack, a small perturbation is the fundamental premise of such models. When designing an adversarial example, the attacker wants the perturbed input to be as close as possible to the original one, in the case of images, close enough that a human can not distinguish one image from the other. We analyze the perturbation concerning scope, limitation, and measurement.

\begin{itemize}[leftmargin=*]
    \item \textbf{Perturbation Scope:} The attacker can generate perturbations that are input specific, in which we call \textit{individual}, or it can generate a single perturbation which will be effective to all inputs in the training dataset, which we call \textit{universal perturbation}.
    
    \item \textbf{Perturbation Limitation:} Two options are possible, \textit{optimized perturbation} and \textit{constraint perturbation}. The optimized perturbation is the goal of the optimization problem, while the constraint perturbation is the set as the constraint to the optimization problem.
    
    \item \textbf{Perturbation Measurement:} Is the metric used to measure the magnitude of the perturbation. The most commonly used metric is the $l_p$-norm, with many algorithms applying $l_0,l_2,l_\infty$ norms.
\end{itemize}

\subsection{Defense Methods}

As seen in \autoref{fig:defenses_adv_attack_generic} based on \cite{xu2019adversarial}, we sub-divide the defenses to adversarial attacks in 3 main categories: Gradient Masking/Obfuscation, Robust Optimization, and Adversarial Example Detection, which are described as:
\begin{itemize}[leftmargin=*]
    \item \textbf{Gradient Masking/Obfuscation:} The core aspect of defense mechanisms based on gradient masking is constructing models with gradients that are not useful for attackers. The gradient masked/obfuscated models, in general, produce loss functions that are very smooth in the neighborhood of the input data. This smoothness around training data points makes it difficult for exploiting algorithms to find meaningful directions towards the generation of an adversarial example.  
    
    \item \textbf{Robust Optimization:} Is a defense strategy that is composed of methods that improve the optimization function either by adding regularization terms, certification bounds, adversarial examples in the objective function, or modifying the model to add uncertainty in the model layers. 
    \item \textbf{Adversarial Example Detection:} Recent work has turned to detect adversarial examples rather than making the DNN robust against creating them. Detecting adversarial examples is usually done by finding statistical outliers or training separate sub-networks that can distinguish between perturbed and normal images.

\end{itemize} 

About the defense mechanisms, we focus this survey on methods related to Robust Optimization. Among the several publications in this survey, each author has its representation and view of robust optimization. In general, even with different notations and representations, most of the papers we have surveyed fit the general representation of Robust Optimization.

The training objective in a DL model is the minimization of the desired loss. The objective is to ajust the model parameters with respect to the labeled data, as seen in \autoref{eq:train_dnn},
\begin{equation}
\label{eq:train_dnn}
    \min_\theta \mathcal{L}(\theta,x,y)
\end{equation}
in which $\theta$ are the model parameters, $x$ is the input to the model, $\mathcal{L}$ is the defined loss function, and $y$ is its true label. With such a formulation, we seek to minimize w.r.t. $\theta$, the loss function. Such formulation fit the parameters to the data points, such that $f(x)$ yields predictions $\hat{y}$ which are equal to the true label $y$. In an adversarial setting this scene changes, in which the objective is different,
\begin{equation}
\label{eq:generate_adv}
    \max_{\delta \leq \Delta} \mathcal{L}(\theta, x+\delta,y)
\end{equation}
in which we are searching for a perturbation $\delta$, smaller than a maximum perturbation $\Delta$, capable of changing the decision of the classifier from prediction $\hat{y}$, to $y'$. The restriction on the perturbation is a designer parameter which is in general defined by the $l_p$-norm.

Equations \ref{eq:train_dnn} and \ref{eq:generate_adv} do not incorporate the data distribution or the restrictions which come from the fact that most of the training datasets do not incorporate the true distribution of the data in which the models will perform inference. Based on the definition from \cite{Madry2018}, we have that, if $\mathbb{D}$ is the true distribution of the data, a training set is draw i.i.d. from $\mathbb{D}$, and is defined as $\mathcal{D} = \{(x_i,y_i)\sim \mathbb{D}\},\text{ for }i=1,...,m$. And the empirical risk of a classifier, which is based on the training set, is defined as:

\begin{equation*}
    R(F, \mathcal{D}) = \frac{1}{|\mathcal{D}|} \sum_{(x,y) \in \mathcal{D}} \mathcal{L}(f(x),y)
\end{equation*}
in which $|\mathcal{D}|$ is the size of the training set $\mathcal{D}$. With that definition the empirical adversarial risk is defined in:
\begin{equation*}
    R_{adv}(F,\mathcal{D}) = \frac{1}{|\mathcal{D}|} \sum_{(x,y) \in \mathcal{D}} \mathcal{L}(f(x+\delta),y)
\end{equation*}

When dealing with adversarial defenses in the lenses of Robust Optimization, the one first solution, is to solve the combined worst-case loss, with the empirical adversarial risk $R_{adv}$, known as adversarial training.

\begin{equation}
    \min_\theta \frac{1}{|\mathcal{D}|} \sum_{(x,y)\in \mathcal{D}} \max_{\delta \in \Delta} \mathcal{L}(f(x+\delta), y)
\label{eq:min_max_adversarial}
\end{equation}

The solution of \autoref{eq:min_max_adversarial}, require special handling or a completely different formulation which define how we categorize the defense mechanisms for adversarial attacks, namely: Adversarial (re)Training, Bayesian Approach, Regularization Approach, and Certified Defenses.

\subsubsection{Adversarial (re)Training as a Defense Mechanism}

The solution of \autoref{eq:min_max_adversarial} requires solving the inner maximization (\autoref{eq:generate_adv}), which is a high dimensional non-convex optimization problem prohibitively hard to solve exactly by standard optimization techniques. The most popular approach to solve such a problem is by approximating \autoref{eq:generate_adv} with the use of a heuristics in which we are searching for a lower bound for \autoref{eq:generate_adv}. While promising and shown to improve robustness even for large models (ImageNet \cite{Xie2019}), such models come with a drawback which when instantiated in practice with the approximation heuristics, they are unable to provide robustness guarantees or certifications. This class of defenses even though very practical to implement can not provide a guarantee that no adversarial example exists in the neighborhood of $x$ capable of fooling $f(.)$.

\subsubsection{Certified Defenses} 

The problem stated in \autoref{eq:min_max_adversarial}, defines the general objective of adversarial training. But as seen, even with the best methods to find a local approximation to the maximization problem, we are subjective to the effectiveness of the attacking method. A way around this inconvenience has been proposed in the literature, which is to exactly solve the maximization problem or approximate to a solvable set of constraints. To formally define Certified Defenses, initially, we consider a threat model where the adversary is allowed to transform an input $x \in \mathbb{R}^d$ into any point from a set $\mathbb{S}_0(x)\subseteq \mathbb{R}^d$. Such set represents the neighborhood of the point $x$ generated by either $l_p$ perturbations, geometric transformations, semantic perturbations, or another kind of transformation in $x$. In case of an $l_p$ perturbation, the set is defined as $\mathbb{S}_0(x) = \{ x' \in \mathbb{R}^d, \mynorm{x-x'}_p < \epsilon \}$.

We further expand the model $f(.)$ as a function of its $k$ hidden layers and parameters $\theta$, where 

\begin{equation}
    f(x) = f^k_\theta \circ f^{k-1}_\theta \circ \dots \circ f^1_\theta
\label{eq:layerwise_nn}
\end{equation}
in which $f^i_\theta : \mathbb{R}^{d_{i-1}}\rightarrow \mathbb{R}^{d_i}$ denotes the nonlinear transformation applied in hidden layer $i$. The objective is to prove a property on the output of the neural network, encoded via a linear constraint:

\begin{equation*}
    c^Tf_\theta(x')+d<0, \forall x' \in \mathbb{S}_0(x)
\end{equation*}
in which $c$ and $d$ are property specific vector and scalar values.

To understand the complexity of the certification, based on \autoref{eq:layerwise_nn}, we define the layer-wise adversarial optimization objective. For $z_1 = x, z_{i+1} = f_i(W_i z_i+b_i)$ :

\begin{maxi}|s|
{z_{1,\dots, d+1}}{(e_y - e_{y_{targ}})^T z_{d+1}}
{}{}
\addConstraint{z_1' \in \mathbb{S}_0}
\addConstraint{z_{i+1} = f_i(W_i z_i+b_i),~ i=1,\dots,d-1}
\addConstraint{z_{d+1} = W_d z_d+b_d}{}
\label{eq:layerwise_optimization}
\end{maxi}
in which $e_i$ unit basis, vectors with value 1 in the class $i^{th}$ position and zeros everywhere else. Such formulation requires special handling given that we have a nonlinear constraint defined by the activation function. 

Several techniques have been proposed to solve such a problem and they are within the scope of study on this survey. The method to train certified neural networks is based on the computation of an upper bound to the inner loss, as opposed to a lower bound computed for adversarial training. These methods are typically referred to as provable defenses as they provide guarantees on the robustness of the resulting network, under any kind of attack inside the threat model. Typical methods to compute the certifications are based on convex relaxations, interval propagation, SMT solvers, abstract interpretation, mixed-integer linear programs, linear relaxations, or combinations of these methods. We explore the diverse techniques in \autoref{sec:certified_defenses}.

\subsubsection{Regularization Approach} 

Regularization techniques focus on making small modifications to the learning algorithm, such that it can generalize better. In a certain way, it improves the performance of the model in unseen data. It prevents model over-fitting to the noise data, by penalizing weight matrices of the nodes. In the specific case of Robust Optimization, the objective of regularization techniques is similar, but focusing on avoiding that small variation on the input, can generate changes in the decision of the algorithm. It does so by either expanding the decision boundaries or limiting changes in the gradient of the model.

Many regularization techniques have been proposed with the most used being the $l_p$ based ones. The $l_2$ regularization technique is introduced to reduce the parameters value which translates to variance reduction. It introduces a penalty term to the original objective function (Loss), adding the weighted sum of the squared parameters of the model. With that, we have a regularized loss $\mathcal{L}_R$ defined as:
\begin{equation*}
    \mathcal{L}_R (x+\delta, y) = \mathcal{L}(x+\delta,y) + \lambda\mynorm{\theta}_2
\end{equation*}
in which a small $\lambda$ lets the parameters to grow unchecked while a large $\lambda$ encourages the reduction of the model parameters. Regularization methods are not restricted to $L_p$ approaches and can involve Lipschitz Regularization, the Jacobian Matrix, and other techniques that we survey on \autoref{sec:regularization_methods}.

\section{Methods for Generating Adversarial Attacks}
\label{sec:adversarial_attacks}
Studying adversarial attacks in the image classification domain improve our insights, as we can visually analyze the dissimilarities between disturbed and non-disturbed inputs. Moreover, the image data, even though high dimensional, are simpler represented than other domains such as audio, graphs, and cyber-security data. Along this section, we'll revise the attack generating algorithms in the image classification domain which can be applied to standard deep neural networks (DNN) and convolutional neural networks (CNN). In which we classify in: white box, black box, and applied real world attacks. In \autoref{tab:adv_att_table} we summarize all the attacks described in this section, highlighting the distance metric used, information access level, algorithm type, and the domain it was applied in the specific publication.

\begin{figure}[!b]
    \centering
    \includegraphics[width=0.9\linewidth]{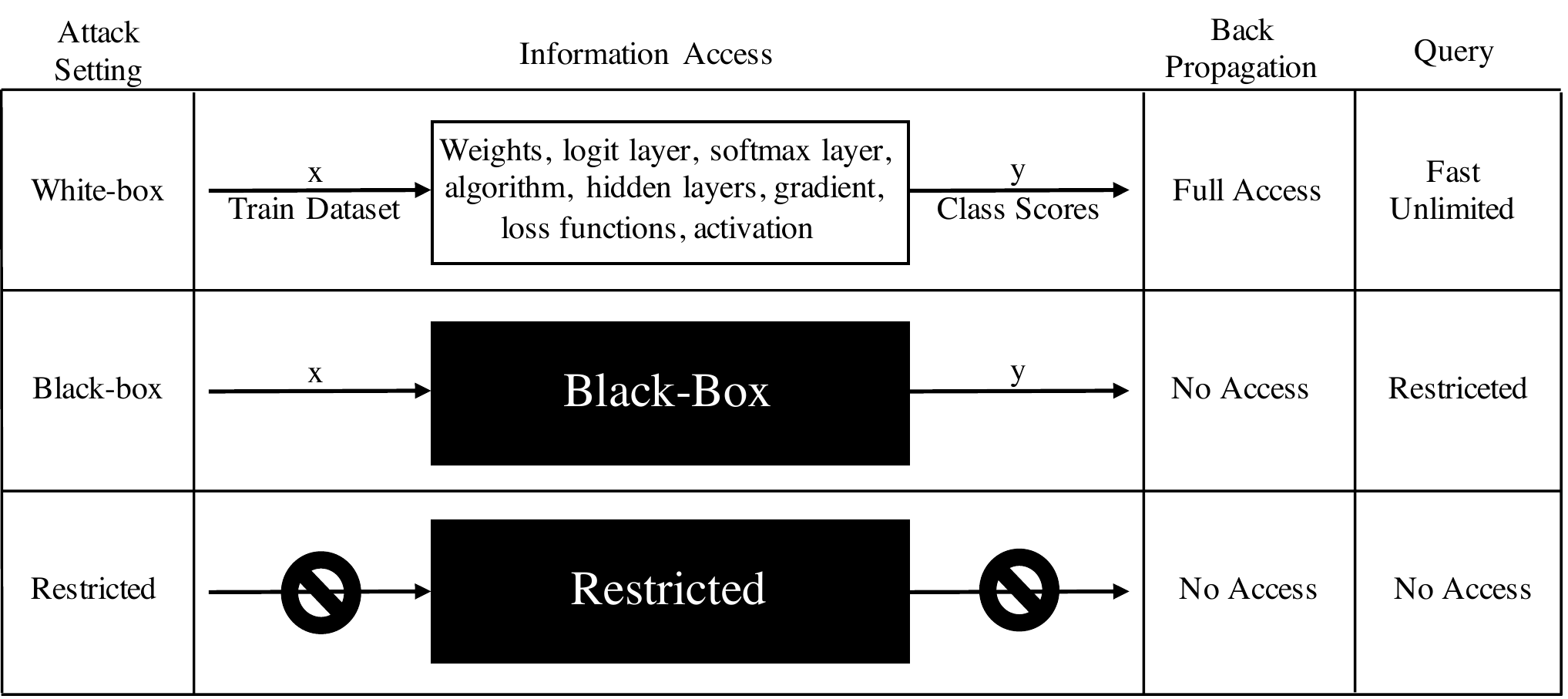}
    \caption{White box and black box attacks diverge mainly on the information the attacker have access to. } 
    \label{fig:attack_definition_wb_bb}
\end{figure}

In the following sub-sections, we list and describe the most popular approaches for the adversarial attack in ML models. We list them in chronological order and focus on giving the most important details on these methods.

\subsection{White-box Attacks}
\label{sec:white_box}

As stated, in white box attacks, there is no restriction in the level of information to which the attacker can access. As a consequence the adversary knows model parameters, dataset, or any other information regarding the model. Under such assumption, given a model $f(.)$, an input $(x,y)$, the main objective is to produce $x'$, which is within certain distance from the original $x$ and maximizes the loss $\mathcal{L}(f(x+\delta),y)$.

\begin{equation*}
    \max_{\delta \in \Delta} \mathcal{L}(f(x+\delta),y)
\end{equation*}

\subsubsection{Box-Constrained L-BFGS}

In, \cite{szegedy2013intriguing}, the existence of small perturbations capable of misleading a classifier were first demonstrated. In the paper, Szegedy et. al. proposed to compute an additive noise $\delta$, which could be added to the original input $x$, capable of misleading the classifier but with minimal or no perceptible distortion to the image. We find the optimal delta, $\delta$, with:

\begin{mini}|s|
{\delta}{c \mynorm{\delta}_2}
{}{}
\addConstraint{f(x+\delta) = y'}
\addConstraint{\text{ all pixel in } (x+\delta) \in [0,1]}
\end{mini}
in which $f(.)$ is the parameterized DNN model, $y$ is the true label, $y'$ is the target label. As is this is a hard problem. Using a box constrained L-BFGS the authors proposed an approximate solution to the problem stated as:

\begin{mini}|s|
{\delta}{c \mynorm{\delta}_2+\mathcal{L}(f(x+\delta),y')}
{}{}
\addConstraint{\text{ all pixel in } (x+\delta) \in [0,1]}
\end{mini}


In this model, Szegedy et. al., managed to generate images that were visually indistinguishable from the original ones, but were able to fool the classifiers into classifying them as another class. This was the first result and publication which has exploited this weakness of deep learning models.  

\subsubsection{Fast Sign Gradient Method}

In , \cite{Goodfellow2015}, introduced a one step adversarial attack framework. The attack image, $x'$, is obtained by a simple additive disturb:
\begin{align*}
    x' =&~ x + \delta_{ut}\\
    x' =&~ x - \delta_{tg}
\end{align*}
in which for the untargeted setting, $\delta_{ut}$, we obtain the perturbation from:
\begin{equation*}
    \max_{\left \| \delta_{ut} \right \|_p\leq \epsilon} \mathcal{L}(f(x+\delta),y)
\end{equation*}
and in the the targeted setting, $\delta_{tg}$, from:
\begin{equation*}
    \max_{\left \| \delta_{tg} \right \|_p\leq \epsilon} (\mathcal{L}(f(x+\delta),y) - \mathcal{L}(f(x+\delta),y'))
\end{equation*}
in which $\epsilon$ is the ball defined normally by an $l_p$-norm.

\begin{figure}
    \centering
    \includegraphics[width=0.7\linewidth]{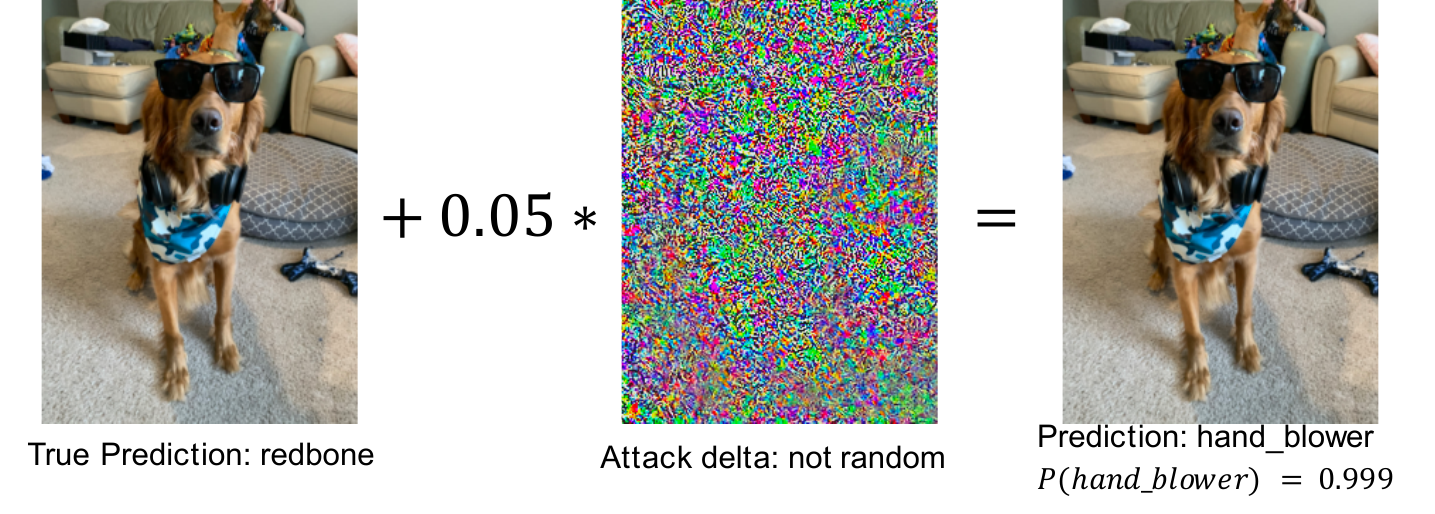}
    \caption{By adding an unoticeble perturbation the dog, previously classified as from the breed redbone, is classified as a Hand blower, with high confidence.}
    \label{fig:FGSM_attack}
\end{figure}

The core of fast sign gradient method maximize the norm of the vector between originally labeled class and the currently assigned label, while in the targeted setting, it focus on minimizing the distance to the target class. As it is a one-step algorithm, it is not very resilient to current defenses but is very fast implementation. \autoref{fig:FGSM_attack} shows the attack of an image and the false prediction.

\subsubsection{DeepFool}

The Deepfool attack proposed by \cite{Moosavi-Dezfooli2016}, is a white box attack which explores the boundaries of the classification model. In the multi-class algorithm, Deepfool initializes with an input $x$ which is assumed to be within the boundaries of the classifier model $f(x)$. With an iterative process, the image is perturbed by a small vector towards the direction of the decision boundaries. The boundaries are approximated by linear functions, more specifically a hyperplane, defined in the algorithm as $\hat{l}$. At each step, the perturbations are accumulated to form the final perturbation to the image. With smaller perturbations than in FGSM (\cite{Goodfellow2015}), the authors have shown similar or better attack success rate. 

\subsubsection{Jacobian-based Saliency Map Attack}

The Jacobian-based Saliency Map attack (JSMA) differs from most of the adversarial attack literature with respect to the norm it uses on the perturbation restriction. While most of the attacks focus on the $l_{\infty}$ or $l_2$ norms, JSMA, proposed in \cite{Papernot2016}, focus on the $l_0$ norm. Under this norm, penalizes the change in a binary way, if the pixel has been changed or not, opposed to $l_2$ based algorithms which takes into consideration the size of the change in the pixels.

In this attack, Papernot et. al., calculates the Jacobian of a score matrix $F$. The model executes the attack in a greedy way. It modifies the pixel which has the highest impact on the model's decision. The Jacobian Matrix is defined as:
\begin{equation*}
    J_F(x) = \frac{\partial F(x)}{\partial(x)} = \{\frac{\partial F_j(x)}{\partial x_i} \}_{x \times j}
\end{equation*}
it models the influence of changes in the input $x$ to the predicted label $\hat{y}$. One at a time pixels from the unperturbed image are modified by the algorithm in order to create a salience map. The main idea of salience map is the correlation between the gradient of the output and the input. It is a guide to the most influential variables of the input, or the ones that probably can deceive the classifier with less manipulation. Based on that, the algorithm performs modifications in the most influential pixel.

\subsubsection{Projected Gradient Descend (PGD)}

Also known as the basic iterative method, was initially proposed in \cite{Kurakin2017}. It is based on the FGSM, but instead a single step of the projected gradient descend, it iterates through more steps, as in:
\begin{align*}
    \delta:=P(\delta+\alpha \nabla_{\delta}\mathcal{L}(f(x+\delta),y))
\end{align*}
in which $P$ denotes the projection over the ball of interest. With such formulation, the PGD, requires more fine-tuning, in choosing the step size $\alpha$. In \cite{Madry2018}, Madry et. al. proposed an iterative method with a random initialization for $\delta$.

\subsubsection{Carlini and Wagner Attacks (CW)}

In \cite{Carlini2017}, 3 $l_p$-norm attacks ($l_0$,$l_2$,$l_{\infty}$) were proposed as a response to \cite{papernot2016distillation}, which proposed the use of Distillation as a defense strategy. In their paper, Papernot et. al., successfully presented a defense mechanism capable of reducing the effectiveness of the FGSM and L-BFGS. CW proposes to solve the same problem stated in FGSM, which is given an input $x$ find a minimal perturbation $\delta$ capable of shifting the classification prediction of the model. The problem is addressed as:
\begin{mini}|s|
{\delta}{c \mynorm{\delta}_p+\mathcal{L}(f(x+\delta),y')}
{}{}
\addConstraint{(x+\delta) \in [0,1]^n}
\end{mini}
in which $\mathcal{L}(f(x+\delta),y')=\max_{i \neq y'}(Z(x')_i)-Z(x')_y)^+$, and $Z(x)=z$ are the logits. As the algorithm minimizes the metrics $\mathcal{L}(.)$, it finds the input $x'$ that has larger score to be classified as $y'$. As we search for the value of $c$, we look for the constant which will produce the smaller distance between $x$ and $x'$.

\subsubsection{Ground Truth Adversarial Example (GTAE)}

So far most of the attacks, even if motivated by the generation of new defenses, are independent of the defense algorithm. In the algorithm proposed by \cite{Carlini2018}, the certified defense proposed in \cite{Katz2017}, is used as a base for the optimization and search for adversarial examples.

The algorithm abstract the $\theta$ and dataset $(x,y)$ with the use of an SMT solver, and solves the system to check if there exist $x'$ near $x$, within the established norm distance, which can cause a misclassification. The ground truth adversarial example is found by reducing the size of $\epsilon$ up to the point that the system can no longer find a suitable $x'$. The adversarial example is considered the $x'$ found with the immediately larger $\epsilon$. It is the first method to calculate an exact provable solution to a minimal perturbation which can fool ML models. In contrast, as stated by the authors, the fact that the model relies on an SMT solver, restrict the applicability of the algorithm to models with no more than a few hundred nodes. This attack has been revisited by \cite{Tjeng2019} and \cite{Xiao2019}.

\subsubsection{Universal Adversarial Perturbations}

Different from the previous methods, the universal adversarial perturbation (UAP), proposed in \cite{moosavi2017universal}, search for a single perturbation capable of fooling all samples from the training dataset. The perturbations, independent of the input, are also restricted to not be detected by humans. The perturbations are constructed based on:
\begin{equation*}
    P_{x \sim \mathcal{D}}(f(x)\neq f(x+\delta))\geq \beta, \text{s. t. } \left \|\delta  \right \|_p \leq \epsilon
\end{equation*}
in which $\epsilon$ defines the size of the perturbation based on an $l_p$-norm and $\beta$ defines the probability of an image sampled from the training dataset being fooled by the generated perturbation. In this case, the algorithm optimizes the probability of fooling the classifier. 

The method to calculate the universal perturbations is based on the DeepFool algorithm, in which the input is gradually pushed towards the model's decision boundary. It differs from DeepFool in the fact that instead of pushing a single input, all members of  $\mathcal{D}$ are modified in the direction of the decision boundary. The perturbations, calculated for each image, are accumulated in a gradual manner. The accumulator is then projected back towards the specified $B_{\epsilon}$ ball, of radius $\epsilon$. Its been shown that with variations of $4\%$, a fooling accuracy of $80\%$ has been achieved.


\subsubsection{Shadow Attack} In \cite{ghiasi2020breaking}, an attack targeting certified defenses was proposed. In their work, they target defenses that certify the model with respect to a radius defined by the $l_p$-norm. One intuitive idea to construct a certified defense is to check within a certain radius $B_\epsilon$ of input, the existence of a perturbation $\delta$, capable of changing the decision of the classifier. The shadow attack is constructed to leverage this premise, and construct a perturbation outside of the certification zone. It is claimed that after labeling an image, these defenses check whether there exists an image of a different label within $\epsilon$ distance (in $l_p$ metric) of the input, where $\epsilon$ is a security parameter selected by the user. If within the $B_\epsilon$ ball all inputs are classified with the same label, then the model is robustly certified. Their model targets not only the classifier but also the certificate. It is done by adding adversarial perturbations to images that are large in the $l_p$-norm and produce attack images that are surrounded by a large ball exclusively containing images of the same label. In order to produce images that are close to the original, in a perception way, but can fool the classifier, they use the following objective function:

\begin{maxi}|s|
{y' \neq y,~\delta}{&- \mathcal{L}(\theta,x+\delta|y') - \lambda_c C(\delta)}
{}{}
\breakObjective{- \lambda_{tv} TV(\delta) - \lambda_s Dissim(\delta)}
\end{maxi}
in which $\mathcal{L}(\theta,x+\delta|\Bar{y})$ refers to the adversarial training loss, $\lambda_c C(\delta)$ is a color regularization term, $\lambda_{tv} TV(\delta)$ is a smoothness penalty term, and $\lambda_s Dissim(\delta)$ guarantees that all color channels receive similar perturbation.

\subsubsection{Other Attacks}

The presented attacks are just part of many more which have been published in many different venues. Here we list some other relevant attack methods available in the literature.
\begin{itemize}
    \item \textbf{EAD: Elastic-net attack} - Similar to L-BFGS the algorithm in \cite{chen2018ead} proposes to find the minimum additive perturbation, which misleads the classifier. Differently it incorporates an association of the norms $l_1$ and $l_2$. It has been shown that strong defenses against $l_\infty$ and $l_2$ norms still fail to reject $l_1$ based attacks.
    \item \textbf{Objective Metrics and Gradient Descend Algorithm (OMGDA)} - The algorithm proposed by \cite{jang2017objective}, is very similar to DeepFool, with the optimization of the step size. Instead of utilizing a fixed and heuristically determined step size in the optimization, in \textit{Jang et al.}, the step size utilizes insights from the softmax layer. The step size is determined based on the size of the desired perturbation and varies over time.
    \item \textbf{Spatially Transformed Attack (STA)} - In \cite{xiao2018spatially}, instead of generating changes in the intensity of the pixels, the authors have proposed a method based on small translational and rotational perturbations. The perturbations are still not noticeable by the human eyes. Similarly in \cite{engstrom2019exploring} the spatial aspect of the input is also exploited for the adversarial example generation.
    
    \item \textbf{Unrestricted Adversarial Examples with Generative Models (UAEGM)} - Based on AC-GAN (\cite{odena2017conditional}), \cite{song2018constructing}, has proposed the use of generative networks to generate examples which are not restricted to being in the neighborhood of the input data. The generated attacks, are not necessarily similar to the ones in the dataset but are similar enough to humans not notice and fool the classifiers.
\end{itemize}  

\subsection{Black-Box Attacks}
Under Black box restriction, models are different from the currently exposed white box, with respect to the information the attacker has access to. In most cases, the adversary does not have any or some information about the targeted model, like the algorithm used, dataset, or parameters, as seen in \autoref{fig:attack_definition_wb_bb}. An important modeling challenge for black-box attacks is to model precisely what information the attacker has about either the learned model or the algorithm. In this sub-section, we list the most relevant methods for black attack generation. 


\subsubsection{Practical Black-Box Attacks (PBBA)}

While assuming access to the information of the model enables a series of attacks, the work in \cite{papernot2017practical}, introduces the possibility of attacking models about which the attacker has less knowledge. In this work, no knowledge about the architecture is assumed, only some idea of the domain of interest. It is also limited the requests for output sent to the model, which requires the attacker to choose wisely the inference requests from the victim's model. To achieve the goal, Papernot et. al. introduce the substitute model framework. The attack strategy is to train a substitute network on a small number of initial queries, then iteratively perturb inputs based on the substitute network’s gradient information to augment the training set. 

\begin{algorithm}
	\begin{algorithmic}[1]
		\caption{Substitute Model}\label{alg:substitute_model}
        \State \textbf{Input}: Substitute dataset $S_0$.
        \State \textbf{Input}: Substitute model architecture $F$.
        \While{$\rho \leq$ epochs}
            \State Label $S_0$ based on queries from original model
            \State Train substitute model based on (3).
            \State Augment dataset based on Jacobian, $S_{\rho} \leftarrow S_{\rho+1}$
        \EndWhile
        \State \textbf{return} 
	\end{algorithmic}
\end{algorithm}

With the boundaries of substitute model adjusted to be close to the original model, any of the methods presented in the previous section can be used to generate the perturbed image.

\subsubsection{Zeroth Order Optimization Based Attack}

In \textit{Chen et al.} \cite{chen2017zoo}, in the attack also known as ZOO, the authors assume accessibility to both, the input data and the confidence scores of the model in which they target their attack. They differ from \cite{papernot2017practical} in the fact that the model does not focus on transferability (creating a substitute model) to achieve the adversarial examples. In their work, they propose a zeroth-order optimization attack which estimates the gradient of the targeted DNN. Instead of traditional gradient descend, they use order 0 coordinate SGD. Moreover, to improve their model and enable the adversarial example generation, they implement dimensionality reduction, hierarchical techniques, and importance sampling. As the pixels are tuned, the algorithm observes the changes in the confidence scores.


Similar to the ZOO, in the one-pixel attack \cite{su2019one}, it is proposed the use of the score confidence to perturb the input and change the decision of the classifier. This paper focuses on modifying a single pixel of the input. With the use of differential evolution, the single pixel is modified in a black-box setting. The authors base the update of the perturbation in the variation of the probability scores for each class. 


\subsubsection{Query-Efficient Black-Box Attacks}

One of the biggest challenges in black-box attacks is the fact that many inference models have mechanisms to restrict the number of queries (when cloud-based or system embedded), or the inference time can restrict the number of queries. One line of research in black-box models looks into making such models more query efficient, for example, the work from \cite{ilyas2017query}, based on natural evolution strategies reduces by 2 or 3 order of magnitude the amount of information requests sent to the model to successfully generate a misclassified perturbed image. The algorithm set queries in the neighborhood of the input $x$. The output of the model is then sampled, and these samples are used to estimate the expectation of the gradient around the point of interest.

\begin{figure}
    \centering
    \includegraphics[width=0.7\linewidth]{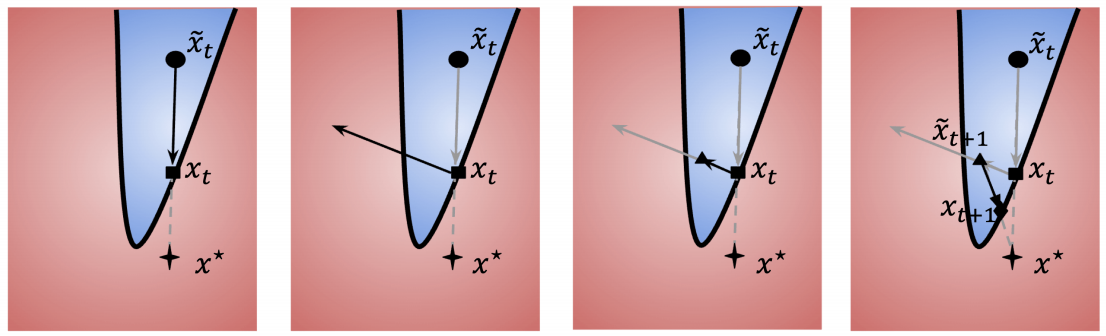}
    \caption{The HopSkipJumpAttack. (a) With binary search find the boundary of the decision. (b) Generate an estimate of the gradient near the decision limit. (c) Update the decision limit point with the use of geometric progression. (d) Do a binary search and update the estimate of the boundary point. Image Source: \cite{chen2019hopskipjumpattack} }
    \label{fig:hopskipjumpattack}
\end{figure}

The algorithm sample the model’s output based on the queries around the input $x$, and estimate the expectation of a gradient of $F$ on $x$. More on the topic, \cite{chen2019hopskipjumpattack}, proposes a family of algorithms based on a new gradient direction estimate using only the binary classification of the model. In their work, it is included $l_\infty$ and $l_2$ norm-based attacks as well as targeted and untargeted attacks. \autoref{fig:hopskipjumpattack} shows an intuition on how the gradient is updated and the boundaries of the decision are used to generate the adversarial attack. Algorithm \ref{alg:bayesian_query} shows how the dimensionality reduction $d^r$ is defined.
Moreover, in the search for query efficient black box attack, \cite{ru2020bayesopt} introduces a method that is independent of the gradient based on Bayesian optimization and Gaussian process surrogate models to find effective adversarial examples. In the model it is assumed that the attacker has no knowledge of the network architecture, weights, gradient or training data of the target model. But it is assumed that the attacker can query model with input $x$, to obtain the prediction scores on all classes $C$. They restrict the perturbation to the $l_\infty$ norm. Their objective is to maximize over the perturbations:
\begin{argmaxi}|s|
{\delta}{[log(f(x_{origin}+g(\delta))_t)}
{}{\delta^*=}
\breakObjective{- log (\sum_{j\neq t}^{C}f(x_{origin}+g(\delta))_j)]}
\addConstraint{\delta \in [-\delta_{max},\delta_{max}]^{d_r}}
\end{argmaxi}
The Bayesian optimization proposed to improve the query efficiency requires the use of a surrogate model to approximate the objective function, in their work a Gaussian Process is used. Moreover to define the next query point is defined by an acquisition function. A big differential in their work is the fact that instead of searching in a high-dimensional space for a perturbation $\delta$, they utilize a function to reduce the dimensionality of the perturbation and later reconstitute to the true image size.

\begin{algorithm}
	\begin{algorithmic}[1]
		\caption{Bayesian selection of $d^r$}\label{alg:bayesian_query}
        \Statex \textbf{Input}: Decoder g(.), observation $\mathcal{D}^d_{t-1}={g(\delta_i), y_i}^{t-1}_{i=1}$ where $g(\delta_i)\in\mathcal{R}^d$ and a set of possible $d^r:{d^r_j}^N_{j=1}$
        \Statex \textbf{Output}: The optimal reduced dimension $d^{r*}$ and corresponding GP model.
        \For{$j=1,...,N$}
            \State $\mathcal{D}^{d^r_j}_{t-1} = \{g^{-1}(g(\delta_i)),y_i\}^{t-1}_{i=1},$ 
            \Comment{$g^{-1}(g(\delta_i)) \in R^{d^r_j}$}
            \State Fit a GP model to $\mathcal{D}^{d^r_j}_{t-1}$ and compute its maximum marginal likelihood $p(\mathcal{D}^d_{t-1}|\theta^*, d^r_j)$
        \EndFor
        \State $d^{r*}=\argmax_{d^r_j\in{d^r_j}^N_{j=1}}p(\mathcal{D}^d_{t-1}|\theta^*, d^r_j)$ and its 
        \State \textbf{return}
	\end{algorithmic}
\end{algorithm}



\subsubsection{Attack on RL algorithm} 
\begin{figure}[!b]
    \centering
    \includegraphics[width=0.7\linewidth]{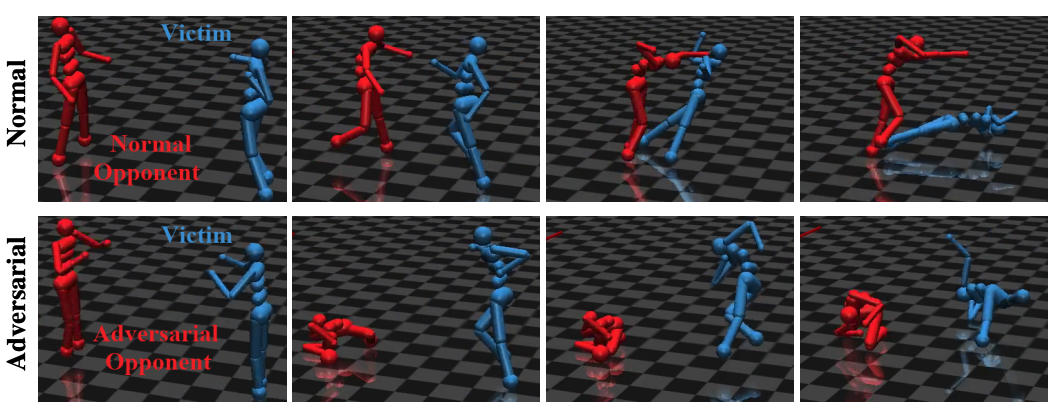}
    \caption{In the first sequence, a strong opponent has to enter collide with the agent to prevent it from winning the game. In the second line, in the adversarial example, the opponent simply crumbles on the floor, which triggers a bad behavior in the victim's policy. Image Source: \cite{gleave2019adversarial}}
    \label{fig:adversarial_policy}
\end{figure}

In \cite{gleave2019adversarial}, a method for generating adversarial examples in reinforcement learning (RL) algorithms was proposed. In RL, an adversarial example can either be a modified image used to capture a state or in the case of this publication, an adversarial policy. It is important to highlight that an adversarial policy is not a strong adversary as we have in two-player games, but one that with a certain behavior triggers a failure in the victim's policy. In this paper, a black-box attack is proposed to trigger bad behaviors in the victim's policy. The victim's policy is trained using Proximal Policy Optimization and learns to "play" against a fair opponent. The adversarial policy is trained to trigger failures in the victim's policy. \autoref{fig:adversarial_policy} shows the difference between an opponent's policy and an adversarial manipulated policy.

Also in the paper, it was shown the dependence of the size of the input space and the effectiveness of adversarial policies. The greater the dimensionality of the observation space under the control of the adversary, the more vulnerable the victim is to attack.

\subsection{Physical World Attack}

The research presented so far is mostly focused on applying the attacks in virtual applications and controlled datasets, but the great concern about the existence of adversarial examples is the extent to which they can imply severe consequences to the users of the system. With that objective, we dedicate this session on exploring publications with real-world applications and consequences clearly stated. 

In \cite{evtimov2017robust}, road signs were physically attacked by placing a sticker in a specific position of the sign. Their attack consisted of initially finding, in a sign image, the location with the most influence for the decision of the classifier. For that objective, an $l_1 norm$ was used because it renders sparse perturbations in the image, making it easier to locate the modification patches. Based on the defined location, an $l_2 norm$ was used to identify the most appropriate color for the sticker. 



Moreover, as face recognition is becoming very popular as a biometric security measure it is the focus of several adversarial attacks. We highlight 4 attacks in face recognition and identification.

\begin{itemize}
    \item \textbf{Evaluation of the robustness of DNN models to Face Recognition against
    adversarial attacks} - In this publication \textit{Goswami et al.} \cite{goswami2018unravelling} evaluates how the depth of the architecture impacts the robustness of the model in identifying faces. They evaluate the robustness with respect to adversarial settings taking into consideration distortions which are normally observed in a common scene. These distortions are handled with ease by shallow networks on the contrary of deep networks. In their approach, they've used Open-Face and VGG-Face networks, and have achieved a high fooling rate. It is important to notice that in their attack no restrictions are made in the visibility of the perturbations.
    \item \textbf{Adversarial Attacks on Face Detectors using Neural Net based Constrained Optimization} In this work, also focusing on prevent face identification, \cite{bose2018adversarial} has generated an attack, based on Carlini and Wagner attack, which was able to fool R-CNN. Their perturbations are not visible in the adversarial example.
    \item \textbf{Generating Adversarial Examples by Makeup Attacks on Face Recognition} - In this research, \cite{zhu2019generating} implement a GAN network to generate make-up perturbation. When the perturbation is applied to the face, the classifier shifts its decision to the target class.
    \item \textbf{Efficient decision-based black-box adversarial attacks on face recognition} - \textit{Dong et al.}\cite{dong2019efficient} propose an evolutionary attack method. The proposed adversarial example generator is constrained in a black-box setting. The algorithm focus on reducing the number of dimensions of the search space by modeling the local geometry of the search vectors. Such an algorithm has been shown to be applicable to most recognition tasks. 
\end{itemize}


\subsubsection{Other Attacks}

In the field of Cyber-security machine learning models are, in general, applied to detect malware, malicious connections, malicious domain classifier, and others. In \textit{Suciu et al.} \cite{suciu2019exploring} an evaluation of the robustness of current malware detection models is performed. The authors retrain the model in a production-scale dataset to perform the evaluation. With the new data, the model which was previously vulnerable to attacks was shown to be stronger and architectural weaknesses were reported. The work of \textit{Suciu et al.} \cite{suciu2019exploring}, explores how attacks transfer in the cyber-security domain, and mainly the inherent trade-off between effectiveness and transferability. With respect to malicious connection and domain, \textit{Chernikova et al.} \cite{chernikova2019adversarial} builds a model that takes into consideration the formal dependencies generated by the normal operations applied in the feature space. The model to generate adversarial examples simultaneously consider both the mathematical dependencies and the real-world constraints of such applications. The algorithm focus on determining the features with higher variability and the ones with a higher correlation with these features. This search is performed for each iteration. All the identified features are modified but constrained to preserve an upper bound on the maximum variation of the features. The upper bound respects physical-world application limitations. 

In the Cyber-Physical domain, several publications demonstrate the brittleness of ML models and the generation of adversarial examples. \cite{melis2017deep} generated adversarial examples to an iCub Humanoid Robot. The attack proposed simply extends over the attacks in \cite{biggio2013evasion}. The main aspect to be considered in this paper is the fact that it highlights the high consequences of the adversarial examples in the decision process of safety-critical applications. Moreover, in self-driving cars several attacks have been derived like DARTS (\cite{sitawarin2018darts}) and the work of \cite{morgulis2019fooling} which shows the attack of traffic signs, the latter with real experiments. In a different sensor type, the work of \cite{cao2019adversarial} demonstrates the attack to a LIDAR sensor, in which they attack the point cloud image.

Moreover in \cite{jia2019fooling}, a novel technique was proposed to attack object tracking algorithms. In their approach, the bounding box is attacked in a single frame, which is enough to fool the algorithm and generate an offset in the placement of the bounding box. Such an attack would be critical to self-driving cars to recognize the position of obstacles, other vehicles, and pedestrians on the road.

    \begin{table}
    \caption{Dichotomy of the Attacks} 
    \centering 
    \begin{tabular}{|c| c| c| c|} %
    \multicolumn{4}{c}{WHITE BOX ATTACKS}  \\
	\hline
    Algorithm & Metric & Step$^a$ & Domain$^b$ \\ 
    \hline 
    \hline
    L-BFGS (\cite{szegedy2013intriguing}) & $l_2$  & Iter. & Im-C \\\hline
    FGSM (\cite{Goodfellow2015}) & $l_\infty, l_2$  & S-Stp & Im-C \\\hline
    Deepfool (\cite{Moosavi-Dezfooli2016}) & $l_2$  & Iter. & Im-C \\\hline
    JSMA (\cite{Papernot2016}) & $l_2$  & Iter. & Im-C \\\hline
    PGD (\cite{Madry2018}) & $l_\infty,l_2$  & Iter. & Im-C \\\hline
    C and W (\cite{Carlini2017}) & $l_0$,$l_2$,$l_{\infty}$  & Iter. & Im-C \\\hline
    GTAE (\cite{Carlini2018}) & $l_0$  & SMT & Im-C \\\hline
    UAP (\cite{moosavi2017universal}) & $l_\infty, l_2$  & Iter. & Im-C \\\hline
    EAD (\cite{chen2018ead}) & $l_1, l_2$  & Iter. & Im-C \\\hline
    OMGDA (\cite{jang2017objective}) & $l_2$  & Iter. & Im-C \\\hline
    STA (\cite{xiao2018spatially}) & Spt-Var  & Iter. & Im-C \\\hline
    UAEGM (\cite{jang2017objective}) & $l_2$  & Iter. & Im-C \\\hline
    Shadow Attack (\cite{ghiasi2020breaking}) & $l_p$  & Iter. & Im-C \\\hline
    RoadSign (\cite{evtimov2017robust}) & $l_1$,$l_2$  & Iter. & S-R \\\hline
    FRA1 (\cite{goswami2018unravelling}) & NR  & S-Stp & F-Rec \\\hline
    FRA2 (\cite{bose2018adversarial}) & $l_2$  & Iter. & F-Rec \\\hline
    FRA3 (\cite{zhu2019generating}) & $l_1$  & Iter. & F-Rec \\\hline
    CSA2 (\cite{chernikova2019adversarial}) & $l_2$  & Iter. & Cyb-Sec \\\hline
    CPA1 (\cite{melis2017deep}) & $l_2$  & Iter. & Cyb-Phy \\\hline
    CPA3 (\cite{morgulis2019fooling}) & $l_0$  & Iter. & Cyb-Phy \\\hline
    CPA4 (\cite{cao2019adversarial}) & NR  & Iter. & Cyb-Phy \\\hline
    CPA5 (\cite{jia2019fooling}) & $l_1,l_2$  & Iter. & Cyb-Phy \\\hline
    \multicolumn{4}{c}{ }  \\ 
	\multicolumn{4}{c}{BLACK BOX ATTACKS}  \\ \hline
	
    Method & Metric & Step$^a$ & Domain$^b$ \\ 
    \hline \hline
	PBBA (\cite{papernot2017practical}) & $l_p$  & Iter. & Im-C \\\hline
    ZOO (\cite{chen2017zoo}) & $l_p$  & Iter. & Im-C \\\hline
    One-Pixel (\cite{su2019one}) & $l_0$  & Iter. & Im-C \\\hline 
    DBA (\cite{brendel2017decision}) & $l_2$  & Iter. & Im-C \\\hline
    HopSkipJumpAttack (\cite{chen2019hopskipjumpattack}) & $l_2$,$l_\infty$  & Iter. & Im-C \\\hline
    UPSET \& ANGRI (\cite{sarkar2017upset}) & $l_\infty$ & Iter. & Im-C \\\hline
    RLAttack (\cite{gleave2019adversarial}) & NR & Iter. & RL \\\hline
    FRA4 (\cite{dong2019efficient}) & $l_2$  & Iter. & F-Rec \\\hline
    CSA1 (\cite{suciu2019exploring}) & $l_\infty$  & Iter. & Cyb-Sec \\\hline
    CPA2 (\cite{sitawarin2018darts}) & $l_1$  & Iter. & Cyb-Phy \\\hline
    \multicolumn{4}{c}{ }\\
    \multicolumn{4}{l}{\footnotesize{\textbf{a} - Iter.: Iterative, S-Stp.: Single Step; \textbf{b} - Im-C.: Image Classification,}}\\
    \multicolumn{4}{l}{\footnotesize{S-R.: Signal Recognition, F-Rec.: Face Recognition, Cyb-Sec.: Cyber-}}\\
    \multicolumn{4}{l}{\footnotesize{Security, Cyb-Phys.: Cyber-Physical, RL.: Reinforcement Learning}}

    \end{tabular}
    \label{tab:adv_att_table}
    \end{table}

\begin{figure}[!t]
    \centering
\begin{tikzpicture}[
  level 1/.style={sibling distance=28mm},
  edge from parent/.style={->,draw},
  >=latex]
   \footnotesize
\centering
\node[root] {Robust Optimization}
  child {node[level 2] (c1) {Adversarial Training}}
  child {node[level 2] (c3) {Certified Defenses}}
  child {node[level 2] (c4) {Regularization Approach}};

\begin{scope}[every node/.style={level 3}]

\node [below of = c1, xshift=8pt] (c12) {\textit{Wang et al. 2020} \cite{wang2020improving}};
\node [below of = c12] (c13) {\textit{Song et al. 2019} \cite{song2019robust}};
\node [below of = c13] (c14) {\textit{Wu et al. 2019} \cite{wu2019defending}};
\node [below of = c14] (c15) {\textit{Wong et al. 2020} \cite{wong2020fast}};
\node [below of = c15] (c16) {\textit{Hu et al. 2020} \cite{hu2020triple}};
\node [below of = c16] (c17) {\textit{Chen et al. 2019} \cite{Chen2019}};
\node [below of = c17] (c18) {\textit{Yang et al. 2019} \cite{Yang2019}};
\node [below of = c18] (c19) {\textit{Kannan et al. 2018} \cite{Kannan2018}};
\node [below of = c19, yshift=-5pt] (c110) {\textit{Matyasko et al. 2018} \cite{Matyasko2018}};
\node [below of = c110] (c111) {\textit{Sinha et al. 2018} \cite{Sinha2018}};
\node [below of = c111] (c112) {\textit{Sen et al. 2020} \cite{sen2020empir}};
\node [below of = c112] (c113) {\textit{Liu et al. 2018} \cite{Liu2018}};
\node [below of = c113] (c114) {\textit{Tramer et al. 2018} \cite{Tramer2018}};
\node [below of = c114] (c115) {\textit{Madry et al. 2018} \cite{Madry2018}};
\node [below of = c115, yshift=-6pt] (c116) {\textit{Goodfellow et al. 2015} \cite{Goodfellow2015}};


\node [below of = c3, xshift=8pt] (c31) {\textit{Wang et al. 2019} \cite{Wang2019}};
\node [below of = c31] (c32) {\textit{Zhai et al. 2020} \cite{zhai2020macer}};
\node [below of = c32, yshift=-5pt] (c33) {\textit{Croce et al. 2019} \cite{croce2019provable}};
\node [below of = c33, yshift=-5pt] (c34) {\textit{Raghunathan et al. 2018} \cite{Raghunathan2018}};
\node [below of = c34, yshift=-10pt] (c35) {\textit{Wong et al. 2018} \cite{Wong2018}};
\node [below of = c35, yshift=-10pt] (c36) {\textit{Boopathy et al. 2019} \cite{boopathy2019cnn}};
\node [below of = c36, yshift=-5pt] (c37) {\textit{Zhang et al. 2018} \cite{Zhang2018}};
\node [below of = c37] (c38) {\textit{Weng et al. 2018} \cite{weng2018towards}};
\node [below of = c38] (c39) {\textit{Weng et al. 2018} \cite{weng2018evaluating}};
\node [below of = c39, yshift=-5pt] (c310) {\textit{Hein et al. 2017} \cite{hein2017formal}};
\node [below of = c310, yshift=-5pt] (c311) {\textit{Singh et al. 2018} \cite{Singh2018}};
\node [below of = c311] (c312) {\textit{Gehr et al. 2018} \cite{Gehr2018}};
\node [below of = c312] (c313) {\textit{Katz et al. 2017} \cite{Katz2017}};

\node [below of = c4, xshift=8pt] (c41) {\textit{Xie et al. 2019} \cite{Xie2019}};
\node [below of = c41] (c42) {\textit{Mao et al. 2019} \cite{mao2019metric}};
\node [below of = c42] (c43) {\textit{Zhang et al. 2019} \cite{Zhang2019}};
\node [below of = c43] (c44) {\textit{Tang et al. 2019} \cite{Tang2019}};
\node [below of = c44] (c45) {\textit{Yan et al. 2018} \cite{Yan2018}};
\node [below of = c45, yshift=-6pt] (c46) {\textit{Ros et al. 2018} \cite{Ros2018}};
\node [below of = c46, yshift=-6pt] (c47) {\textit{Cisse et al. 2017} \cite{Cisse2017}};
\node [below of = c47] (c48) {\textit{Sokolic et al. 2017} \cite{Sokolic2017}};
\node [below of = c48, yshift=-6pt] (c49) {\textit{Gu et al. 2015} \cite{Gu2015}};

\end{scope}

\foreach \value in {2,...,16}
  \draw[->] (c1.190) |- (c1\value.west);


\foreach \value in {1,...,13}
  \draw[->] (c3.190) |- (c3\value.west);

\foreach \value in {1,...,9}
  \draw[->] (c4.190) |- (c4\value.west);

\end{tikzpicture}
\caption{Summary of the defense mechanisms sub-divided in the 3 categories, namely: Adversarial Training, Certified Defenses, and Regularization Approach.}
\label{fig:defenses_complete}
\end{figure}
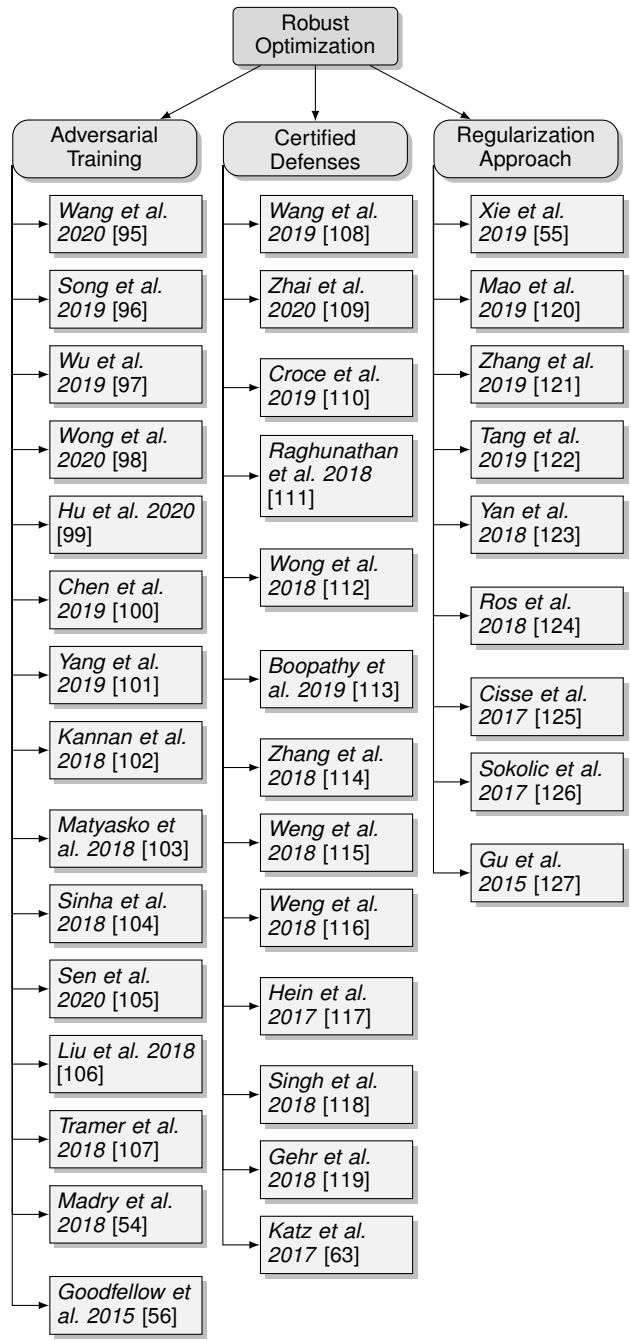

\section{Defense Mechanisms based on Robust Optimization Against Adversarial Attacks}
\label{sec:robust_optimization}
From the examples presented so far, we see that both DNN's and CNN's are very unstable locally and susceptible to misclassify examples with perturbations that are barely perceivable by the human eye. Several works have reported structured algorithms and formulations to improve the robustness of ML models employing robust optimization. The goal of this section is to visit the most commonly identified techniques to achieve robustness (through the eyes of the optimization), namely Adversarial Training, Bayesian Approach, Certified Defenses, and Regularization Approaches, which are summarized in \autoref{fig:defenses_complete}.

We summarize in \autoref{tab:results_adversarial_training} the results presented in the surveyed papers. We've compiled the information available in each of those papers concerning the error rate under the tested conditions. Each of the papers has different evaluation criteria and conditions. The dataset is highly influential in the accuracy of the model. To that end, we've been careful to better express the conditions in which the models were evaluated. Papers in which the results were not clear or have used a very specific metric are not listed in these tables to keep consistency among the results.

\subsection{Defending through Adversarial (re)Training}
\label{sec:adversarial_training}

After the first introduction of adversarial examples (\cite{szegedy2013intriguing} and \cite{biggio2013evasion}), defense mechanisms to train robust neural networks were built based on the inclusion of adversarial examples to the training set. Models trained using adversarial training with projected gradient descent (PGD) were shown to be robust against the strongest known attacks. This is in contrast to other defense mechanisms that have been broken by new attack techniques. In this section, we explore robustness mechanisms that explicitly or implicitly address the robustness in deep learning through either adding adversarial examples in the dataset or incorporate them in the objective function for the optimization.

\subsubsection{Harnessing Adversarial examples} In his work, \cite{Goodfellow2015} Goodfellow et. al. suggests the use of adversarial examples in the training process to improve the robustness of machine learning models. It is a very simple idea, which worked for the proposed configuration. The algorithm, using FGSM in an untargeted setting, would generate a set of adversarial examples $x'$, which were fed to the learning algorithm with the true label, $(x',y)$. Important to notice the limitation of such a framework, it is robust against FGSM attacks, but susceptible to other attacks, such as iterative methods. Such weakness was pointed in a later work by \cite{Tramer2018}, which also shown single-step attacks could fool such a defense.

In \cite{wong2020fast}, it is studied an adaptation of FGSM in adversarial training. The initially proposed FGSM training was shown to create a massive over-fitting in the model having it not robust to iterative attack methods such as PGD. In this new publication, Wong et. al., proposes small modifications in the initialization of the FGSM algorithm to accommodate randomness as a way to prevent over-fitting in the training process. Instead of having a fixed initialization, the perturbations are generated as follows:
\begin{enumerate}
    \item $\delta = Uniform(-\epsilon, \epsilon)$
    \item $\delta = \delta + \alpha sign(\nabla_\delta \mathcal{L}(f_\theta(x_i+\delta),y_i))$
    \item $\delta = \max(\min(\delta,\epsilon),-\epsilon)$
\end{enumerate}
The addition of the sampling from a uniform distribution to the initial perturbation allowed the algorithm to better estimate the inner maximization, improving the robustness of adversarially trained models with FGSM, maintaining the speed of the adversarial example generation.

\subsubsection{Towards Deep Learning Models Resistant to Adversarial Examples} In their publication, Madry et. al. \cite{Madry2018}, among an extensive discussion of robustness in the machine learning models, propose to incorporate iterative methods to approximate the inner maximization problem shown in:

\begin{equation}
    \label{eq:min_max_madry}
    \min_{\theta}( \frac{1}{\mathcal{D}} \sum_{(x,y)\in~\mathcal{D}} \max_{\delta \in \Delta(x)} \mathcal{L}(f(x+\delta),y))
\end{equation}
where $\delta$ is the perturbation, $f(.)$ is the model. 

In this publication, the authors have approximated the inner maximization problem with the PGD attack method, but it is important to highlight in this formulation, that the model will be as robust as the attack it was trained on. If new and more efficient adversarial examples are presented to the model at inference time, nothing can be stated regarding the robustness of the model.

\subsubsection{Ensemble Adversarial Training} In their research \cite{Tramer2018}, the authors show that when the model is trained directly on single/iterative methods, the model trains on examples crafted to maximize a linear approximation of the loss. Moreover, such a model training method converges to a degenerate global minimum. These artifacts near the inputs $x$ obfuscate the approximation of the loss. According to their findings, the model learns weak perturbations rather than generating robustness against strong ones.

As a countermeasure, the paper implements a framework in which the model is trained with adversarial examples from similar classifiers. In their method, the generation of adversarial examples is dissociated from the network being optimized. Under such circumstances, their proposed framework approximates to the work in which black-box attacks are generated with the use of auxiliary models, and consequently makes their approach more resilient to it. In the algorithm, to train classifier $F_0$, they train other sets of classifiers, $F_1, F_2,..., F_n$, generate adversarial examples for these classifiers, and use the adversarially generated examples to train $F_0$. Their defense model even for the ImageNet dataset, has shown robustness against black-box attacks.

In a different work, using a similar concept, \cite{Liu2018} proposes the augmentation of the model robustness with the use of random noise layers to prevent the strong gradient attacks. It is a gradient masking attack, but the authors position the algorithm as an improvement to ensemble adversarial defense, by stating that the defense is equivalent to ensembling an infinite number of noisy models. Such ensembling is claimed by the authors to be equivalent to training the original model with a Lipschitz regularizing term. They've achieved significant results against Carlini and Wagner's strong attacks. Algorithm \ref{alg:rand_self_ensemble} shows the steps to implement the proposed algorithm.

\begin{algorithm}
	\begin{algorithmic}[1]
		\caption{Random Self-Ensemble}\label{alg:rand_self_ensemble}
		\Statex Training
        \For{iter = 1,2,...}
            \State Randomly sample $(x_i,y_i)$ in dataset
            \State Randomly generate $\epsilon \sim \mathcal{N}(0,\sigma^2) $ for each noisy layer
            \State Compute the noise gradient, $g = \nabla_\theta \mathcal{L}(f_\epsilon(\theta,x_i),y_i)$
            \State Update weights: $\theta' \leftarrow \theta - g$
        \EndFor
        \Statex Testing
        Given a test image $x$, initialize $p=(0,...,0)$
        \For{$j={1,2,...,Ensembles}$}
            \State Generate $\epsilon$
            \State Forward propagation and generate probability output, $p^j$
            \State Update, $p \leftarrow p+p^j$
        \EndFor
        \State \textbf{Return class with with maximum probability} 
	\end{algorithmic}
\end{algorithm}

More recently in \cite{sen2020empir}, a mixed-precision ensemble was proposed. Ensemble of Mixed Precision Deep Networks for Increased Robustness (EMPIR) is based on the observation that quantized neural networks often show higher robustness to adversarial attacks than full precision networks. Such models sacrifice accuracy. EMPIR combines the accuracy of full models with the robustness of quantized models by composing them in an ensemble.

\subsubsection{Principled Adversarial Training} In \cite{Sinha2018}, Sinha et. al. have presented a method to solve the outer maximization with a Lagrange relaxation approach. By doing so, they presented both, an adversarial training method based on the min-max formulation and a method to prove the robustness of the model. In addition to the adversarial training, a penalty is added on the loss term to help regularize the learning. It is used as a Lagrangian formulation to generate this penalty. It perturbs the underlying data distribution within a Wasserstein ball. The model's efficiency is restricted to smooth losses, but under such constraint, it can achieve a moderate level of robustness. The computational or statistical cost, when compared to empirical risk minimization is small. 

\subsubsection{Network Robustness with Adversary Critic}   The Adversarial Training formulation is an optimization problem that naturally involves minimization and maximization. In \cite{Matyasko2018} and \cite{weng2018evaluating} the GAN framework is proposed to generate the noisy perturbations, which will lead to the adversarial example. On the other-hand the discriminator of the network act as a critic which will discern if the presented input $x$ is adversarial or not. In the work, it is highlighted that the generated adversarial networks are also robust to black-box attacks, showing similar or better performance than state-of-the-art defense mechanisms.  


\subsubsection{Adversarial Logit Pairing}

In \cite{Madry2018} Madry et. al. suggested the use of the \autoref{eq:min_max_madry} to adversarial train the model, and achieve robustness in their model. \textit{Kannan et al.}\cite{Kannan2018} implements a mixed version of this defense. Instead of training the robust model only on adversary perturbed images, they incorporate a mix of clean $(x,y)$ and perturbed $(x',y)$ batches, which they call mixed mini-batch PGD (M-PGD). 

In their work, they go beyond to analyze the fact that most of the adversarial training framework, train the models with information that $x'$ should belong to the class $t$, but the model is not given any information indicating that $x'$ is more similar to the actual sample $x$ than any other belonging to the same class. To that extent, they propose another algorithm called adversarial logit pairing.

For a model $f(.)$ trained on a mini-batch $\Gamma$ of clean examples $\{x_1, x_2,...,x_m\}$ and corresponding adversarial examples $\{x'_1, x'_2, ..., x'_m\}$, with $f(x)$ mapping the input to a prediction. With $\mathcal{L}(\Gamma,\theta)$ being the cost function used to perform the adversarial training, the adversarial logit pairing consists of minimizing the loss:

\begin{equation*}
    \mathcal{L}(\Gamma,\theta)+\lambda \frac{1}{m} \sum_{i=1}^m \mathcal{L}(f(x_i),f(x'_i))
\end{equation*}

\subsubsection{ME-Net} ME-Net \cite{Yang2019} introduces the concept of utilizing matrix estimation as a way to augment the adversarial sample size and eliminate the perturbations from adversarial examples. 


The training procedure to ME-Net is described as follows, the algorithm creates a mask in which each pixel is preserved with probability $p$, and set to zero with probability $1-p$. For each input image $x$, $n$ masks are applied, with different pixel drop probability. The generated masked images, $X$, are then processed through a Matrix Estimation algorithm, which would obtain the reconstructed images $\hat{X}$. The DNN model is trained on the reconstructed images, which can be further processed with the use of more adversarial training techniques. For inference, to each input $x$, a mask is randomly sampled from the pool of masks obtained in the training time, applied to $x$, and then reconstruct to generate $\hat{x}$. The process of masking and reconstructing the images is claimed by the authors to reduce the effects of adversarial perturbations in the image.



\subsubsection{Robust Dynamic Inference Networks} In \cite{hu2020triple}, an input-adaptive dynamic inference model to the adversarial defense is proposed. In this method each input, regardless of clean or adversarial samples, adaptively chooses which output layer to take for its prediction. Therefore, a large portion of input inferences can be terminated early when the samples can already be inferred with high confidence. The benefit of the use of such models comes from the fact that the multiple sources of losses provide much larger flexibility to compose attacks (and defenses), compared to the typical framework. In this work, a methodology to attack and defend in such models is proposed.


\subsubsection{Defending against Occlusion Attacks} In \cite{wu2019defending}, the authors investigate defenses against physically realizable attacks, more specifically they investigate defenses to attacks in which part of the object is occluded by a physical patch. It's been demonstrated that adversarial training with either PGD or Randomized Smoothing did not improve the robustness of the models significantly. In their work, they propose, implement, and use a Rectangular Occlusion Attack(ROA). The ROA enabled the emulation of physical attacks in the virtual world and the consequent training of adversarial resistant models. 

\subsubsection{Robust Local Features for Improving the generalization of Adversarial Training} In their research,\textit{Song et al.} \cite{song2019robust} investigates the fact that when models are trained with no adversarial techniques they gather better information about local features, which improves the model's generalization ability. Opposed, DNN models which were adversarially trained tend to have a bias in favor of a global understanding of the features.  In their work, they propose a method to train adversarial robust models, which are biased towards local features.  In their work they define the Random Block Shuffle, to randomize the features of input inside the image. Such an approach prevents the adversarial learning method from learning only global features. The model learns from a combination of shuffled and unshuffled images.
\begin{align}
    \label{eq:loss_feature_bias}
    \begin{split}
    \mathcal{L}_{RLFAT_P}(F;x,y) =& \mathcal{L}_{PGDAT}^{RLFL}(F;x,y) +\\
    &\eta \mathcal{L}_{PGDAT}^{RLFT}(F;x,y) \\
    \mathcal{L}_{RLFAT_T}(F;x,y) =& \mathcal{L}_{TRADES}^{RLFL}(F;x,y) +\\ 
    &\eta \mathcal{L}_{TRADES}^{RLFT}(F;x,y)
    \end{split}
\end{align}
In the loss defined in \autoref{eq:loss_feature_bias}, the factor $\eta$ balances the contribution between the local feature-oriented loss and the global oriented.

\subsubsection{Misclassification Aware Adversarial Training} In \cite{wang2020improving}, the authors propose the analysis of the misclassification examples, intending to improve the accuracy of the model against perturbed inputs. 
To perform such a study they've trained a classifier with 10-step PGD, obtaining 87\% training accuracy, they extracted the 13\% misclassified examples and sampled 13\% correctly classified examples from the training dataset. The examples originally misclassified by the model, are the ones that impact the most the final robustness. Compared to standard adversarial training the final robustness drops drastically if misclassified examples are not perturbed during adversarial training. In contrast, the same operation on sampled correctly classified examples only slightly affects the final robustness.

Based on the observations a regularization term is proposed to incorporate an explicit differentiation of misclassified examples. Initially they propose a Boosted Cross Entropy loss, defined as:
\begin{align*}
    \text{BCE} (p(\hat{x}_i^{\prime},\theta),y_i) =& -log(p_{y_i}(\hat{x}_i^{\prime},\theta))\\&-log(1-\max_{k\neq y_i} p_t(\hat{x}_i^{\prime},\theta))
\end{align*}
in which $p_t(\hat{x}_i^{\prime},\theta)$ is the softmax on logits of $x_i$ belonging to class $t$. With that the objective function is defined as:
\begin{align*}
    \mathcal{L}^{\text{MART}}(\theta) &= \frac{1}{n}\sum^n_{i=1}\mathcal{L}(x_i,y_i,\theta)\\
    \mathcal{L}(x_i,y_i,\theta)&:= BCE(p(\hat{x}_i^{\prime},\theta),y_i))\\ &+ \lambda \text{KL}(p(x_i, \theta)||p(\hat{x}_i^{\prime},\theta),y_i))(1-p_{y_i}(x_i,\theta))
\end{align*}

\subsubsection{Input Transformation Methods}

Input transformation methods propose to train the network in transformed images, such as bit-depth reduction, JPEG compression, total variance minimization, and image
quilting or an ensemble of these methods to improve the robustness of the model.In \cite{das2018shield}, the use of JPEG compression was proposed as a countermeasure of the pixel displacement generated by the adversarial attacks. In \cite{guo2018countering}, a combination of total variation minimization and image quilting is used to defend against strong attacks. Even-though these transformations are nonlinear, a neural network was used to approximate the transformations, making them differentiable, and consequently easier to obtain the gradient. Different in \cite{raff2019barrage}, an ensemble of weak transformation defenses was proposed to improve the robustness of the models, among the transformations included in the defense are: color precision reduction, JPEG noise, Swirl, Noise Injection, FFT perturbation, Zoom Group, Color Space Group, Contrast Group, Grey Scale Group, and Denoising Group.

\begin{table*}
\caption{Results of defenses based on Adversarial (re)Training and Regularization} 
\centering 
\begin{tabularx}{0.895\linewidth}{ | c | c | c | c | c | c | c | }
	\multicolumn{7}{c}{ADVERSARIAL (RE)TRAINING}  \\
	\hline
	Publication & Architecture & Dataset & Norm & Adversarial & $\epsilon$ & Error Rate \\ 
	\hline
	\textit{Goodfellow et al. 2015} \cite{Goodfellow2015} & DNN & MNIST & $l_2$ & L-BFGS & 0.25 & 17.9\% \\ \hline
	\textit{Tramer et al. 2018} \cite{Tramer2018} & Inception ResNet v2 & ImageNet & $l_\infty$ & Step-LL & 16/256 & 7.9\% \\ \hline
	\textit{Madry et al. 2018} \cite{Madry2018} & ResNet & CIFAR-10 & $l_2$ & PGD & 8 & 54.2\% \\ \hline
	\textit{Liu el al. 2018} \cite{Liu2018} & VGG16 & CIFAR-10 & $l_\infty$ & C\&W & 8/256 & 10\% \\ \hline
	\textit{Sen et al. 2020 }\cite{sen2020empir} & AlexNet & ImageNet & $l_\infty$ & C\&W & 0.3 & 70.64\% \\ \hline
	\textit{Kannan et al. 2018} \cite{Kannan2018} & ResNet-101 & ImageNet & $l_2$ & PGD & 12/255 & 55.60\% \\ \hline
    \textit{Hu et al. 2020} \cite{hu2020triple} & ResNet38 & CIFAR-10 & $l_2$ & PGD & 8/255 & 30.29\% \\ \hline
	\textit{Wong et al. 2020} \cite{wong2020fast} & PreAct ResNet18 & ImageNet & $l_\infty$ & R-FGSM & 2/255 & 56.7\% \\ \hline
	\textit{Song et al. 2019} \cite{song2019robust} & ResNet w32-10 & CIFAR-100 & $l_\infty$ & PGD & 0.03 & 68.01\% \\ \hline
	\textit{Wang et al. 2020} \cite{wang2020improving} & ResNet-18 & CIFAR-10 & $l_\infty$ & PGD & 8/255 & 45.13\% \\ \hline
	\multicolumn{7}{c}{ }  \\ 
	\multicolumn{7}{c}{REGULARIZATION APPROACH}  \\ \hline
	Publication & Architecture & Dataset & Norm & Adversarial & $\epsilon$ & Error Rate \\ \hline
	\textit{Cisse et al. 2017} \cite{Cisse2017} & ResNet & CIFAR-100 & SNR($x$,$\delta$), $l_\infty$ & FGSM & 33 & 47.4\% \\ \hline
	\textit{Yan et al. 2018} \cite{Yan2018} & ResNet & ImageNet & $l_2$ & FGSM & 2.43E-3 & 50\% \\ \hline
	\textit{Zhang et al. 2019} \cite{Zhang2019} & ResNet & CIFAR-10 & $l_\infty$ & C\&W & 3.1E-2 & 18.76\% \\ \hline
	\textit{Xie et al. 2019} \cite{Xie2019} & ResNet-152 & ImageNet & $l_\infty$ & PGD & 16 & 57.4\% \\ \hline
	\textit{Mao et al. 2019} \cite{mao2019metric} & Modified LeNet & Tiny ImageNet & $l_\infty$ & C\&W & 8/255 & 82.52\% \\ \hline
	\textit{Shafahi et al. 2019} \cite{shafahi2019adversarially} & Wide-ResNet-32 & CIFAR-100+$\rightarrow$10+ & $l_2$ & PGD & 8 & 82.3\% \\ \hline 
\end{tabularx}
\label{tab:results_adversarial_training}
\end{table*}

\subsection{Regularization Techniques}
\label{sec:regularization_methods}

As shown in \autoref{sec:white_box}, several algorithms depend on the model's gradient to estimate the local optima perturbations which will fool the classifier. A stream of research towards adversarial robust optimization, look into applying regularization approaches to reduce the influence of small perturbations in the input on the output decisions. In this section, we review some relevant publications in the field in which the author's objective is to improve the robustness of the model.

\subsubsection{Towards DNN Architectures Robust to Adversarial Examples} In their work, Gu et. al. \cite{Gu2015} propose the use of Contractive Autoencoder. To regularize the gradient, they add a penalty to the loss in the back-propagation concerning the partial derivatives at each layer. By incorporating a layer-wise contractive penalty (in partial derivatives), they show that adversarial generated from such networks have significantly higher distortion. In their approach, the network could still be fooled by adversarial examples, but the level of noise necessary to fool such a network is considerably higher than standard networks in which there is no contractive penalty.


\subsubsection{Robust Large Margin Deep Neural Networks} The work presented in \cite{Sokolic2017} analyzes the generalization error of DNN's through their classification margin. In their work, they initially derive bounds to the Generalization Error (GE) (adversarial attacks as consequence) and express these bounds as a dependence of the model's Jacobian Matrix (JM). In their work, it was shown that the depth of the architecture does not affect the existence of a GE bound, conditioned to fact that the spectral norm of the JM is also bounded, around the training inputs. With this definition a weigh and batch normalization regularizer is derived. The regularizer is based on the bound derived based on the JM.

\subsubsection{Input Gradient Regularization} Parseval networks \cite{Cisse2017} is a layer-wise regularization method to reduce the networks variability to small disturbances in the input $x$. The work starts with the principle that DNN's are a composition of functions presented by layers. To keep the variability of the output controlled, they propose to maintain the Lipschitz constant small at every hidden layer (for fully connected, convolutional, or residual layer). For that, they analyze the spectral norm of the weight matrix.

In \cite{Ros2018}, Ros et. al. based on the same principle of gradient regularization, propose the use of such techniques to improve both the robustness and interpretability of Deep Neural Networks. The authors claim that raw input gradients are what many attacks use to generate adversarial examples. Explanation techniques that smooth out gradients in background pixels may be inappropriately hiding the fact that the model is quite sensitive to them. They hypothesize that by training a model to have smooth input gradients with fewer extreme values, it would not only make the model more interpretable but also more resistant to adversarial examples. Their gradient regularization is given by:
\begin{align*}
    \theta^* &= \argmin_{\theta} \sum^N_{n=1} \sum^K_{k=1} -y_{nk}log(f_\theta(X_n)_k)\\
    &+\lambda \sum^D_{d=1} \sum^N_{n=1} (\frac{\partial}{\partial x_d} \sum_{k=1}^K -y_{nk}log(f_\theta(X_n)_k))^2
\end{align*}
in which $\lambda$ specifies the penalty strength. The goal of this update is to ensure that if any input changes slightly the KL divergence between predictions will not change significantly.

\subsubsection{DeepDefense} \textit{Yan et al.}\cite{Yan2018} propose the algorithm called DeepDefense focusing on the improvement of the robustness of the DNN models, which is an regularization method based on the generated adversarial perturbation. Similar to adversarial (re)Training the algorithm incorporates the adversarial generation the loss function. It does not occur as shown in \autoref{eq:min_max_adversarial}, differently it is presented as:
\begin{align*}
    \min_\theta \sum_k \mathcal{L}(x_k, y_k, \theta) + \lambda \sum_k R(-\frac{\left \|\delta_{x_k}  \right \|_p}{\left \| x_k \right \|_p})
\end{align*}
in which we see that at the same time that the loss is minimized a regularization term based on the perturbation $\delta_{x_k}$ is added to penalize the norm of the adversarial perturbations. The penalty function $R(.)$, treats the samples different depoending if they were correctly classified or not, it increases monotonically when the sample is correctly classified. With such a behavior the function gives preference to those parameter settings which are able to resist even to small $\frac{\left \|\delta_{x_k}  \right \|_p}{\left \| x_k \right \|_p}$.

\subsubsection{TRADES} 
\begin{figure}[!b]
    \centering
    \includegraphics[width=0.6\linewidth]{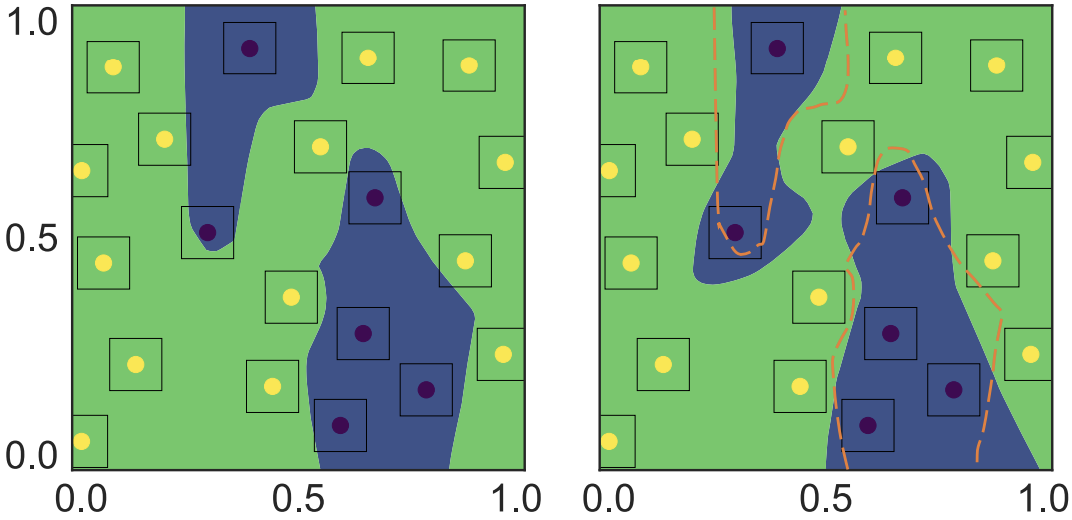}
    \caption{In the \textbf{left} we see the boundaries of a model trained with standard DNN. In the \textbf{right} the boundaries of the decision are pushed further from the data points, showing not so sharp transitions compared to naturally trained methods. Image Source: \cite{Zhang2019}}
    \label{fig:trades_reg}
\end{figure}

In their work, Zhang et. al. \cite{Zhang2019} states the intrinsic trade-off between robustness and accuracy. In their work, they derive a differentiable upper bound for the natural and boundary errors of the DNN model. To derive such bound, the error generated by the adversarial examples (called robust error) is decomposed in two parts: 1 - the natural misclassification, and 2 - the boundary errors. The bounds are shown to be the tightest overall probability distributions. Based on these bounds a defense mechanism called TRADES is proposed. In its core, the algorithm still minimizes the natural loss, consequently increasing the accuracy of the model, but at the same time, it introduces a regularization term, which induces a shift of the decision boundaries away from the training data points. The expansion of the decision boundaries can be seen in \autoref{fig:trades_reg}.

\subsubsection{Metric Learning for adversarial Robustness} \cite{mao2019metric} focuses on learning a distance metric for the latent representation of the input. Through an empirical analysis, the authors have observed that inputs under PGD adversarial attacks shift its latent representation to a false class. The shift in the latent representations spread in the false class and become indistinguishable from the original images in the class. In their paper, they've added a new constraint to the model with metric learning. Their model implements a variation of the naive triplet loss, called TLA (triplet loss adversarial training), which overcome the variance of the adversarial data over the false class. The TLA works by approximating the samples from the same class, independently if they are adversarial or unperturbed samples, and enlarge the boundary distance concerning other classes.

\subsubsection{Adversarially Robust Transfer Learning}  In \cite{shafahi2019adversarially}, a study on transfer learning for robust models is performed. In their paper, they robustly train a Wide-ResNet 32-10 \cite{zagoruyko2016paying} using the algorithm proposed in \cite{Madry2018}. In their first experiment, they sub-divide the model in blocks of layers and analyze the impact of changing the number of unfrozen blocks in the transfer learning process. It was seen that when only the fully connected layer and the batch normalization blocks were re-trained, the network had similar or improved robustness in the new domain. Opposed, when more blocks were re-trained the accuracy and robustness dropped drastically. The authors claimed that the robust model's feature extractors act as filters that ignore the irrelevant parts of the images. 

With the intent of improving the overall performance of classifiers transferred from a robust source model by improving their generalization on natural images, the authors proposed an end-to-end transfer learning model with a no-forgetting property. To do so, they only fine-tune the feature extraction parameters $\theta$. It consists basically of the addition of a regularization term to the loss of the model. 
\begin{equation*}
    \min_{\theta,w} \mathcal{L}(f(x,\theta),y,w) + \lambda l_p(f(x,\theta),f_0(x,\theta^*))
\end{equation*}

\subsection{Certified Defenses}
\label{sec:certified_defenses}

Certified defenses try to theoretically find certificates in distances or probability to certify the robustness of DNN models. These methods are explored in this section.

\subsubsection{Exact Methods} The work of \cite{Katz2017}, was a big first step in the direction of formalizing methods to certify the robustness of neural networks. In their work, also, to provide a formulation for SMT solver for the ReLU activation function, they've proven safety in a small neural network for aircraft collision prediction. They were able to use their solver to prove/disprove local adversarial robustness for their example DNN for a few arbitrary combinations of input $x$ and disturbances $\delta$. In their experiments, the verification took from few seconds to a few hours depending on the size of the perturbation, with bigger perturbations taking longer to be verified. 

\begin{figure}[!ht]
    \centering
    \includegraphics[width=0.7\linewidth]{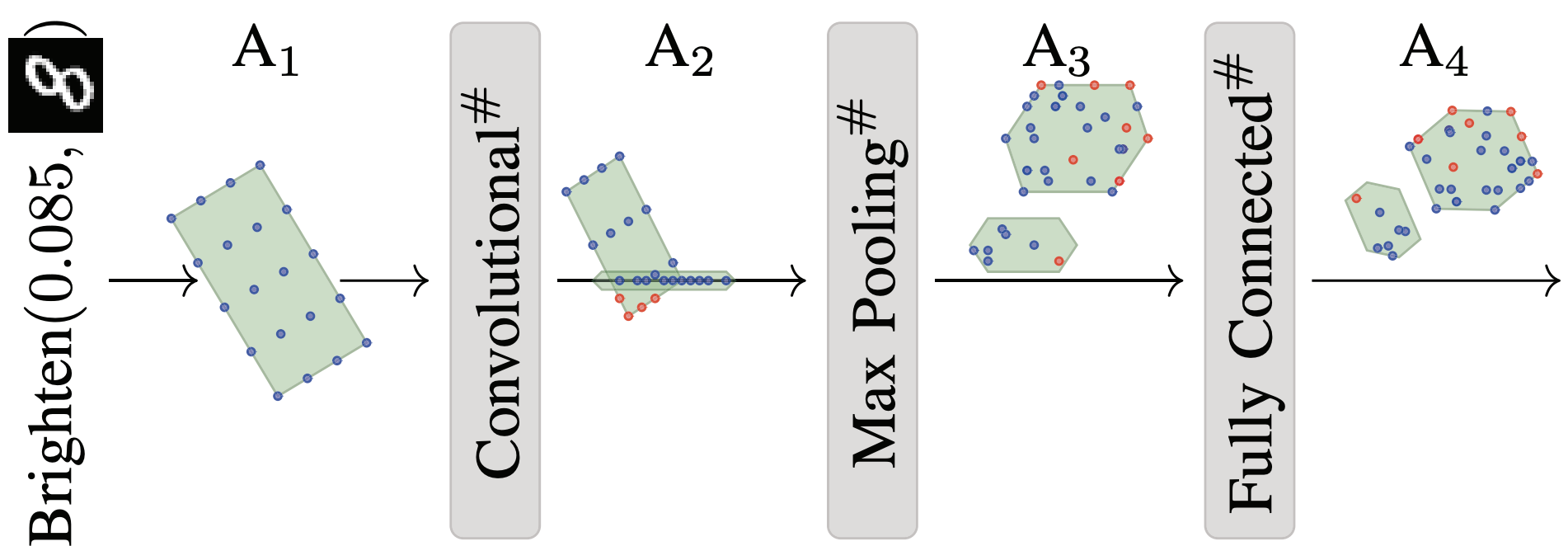}
    \caption{From left to right, initially the algorithm generates an abstract element which encompass all perturbed images. It propagates through all the abstractions of the layers. The verification is successful if all images are in the same classification group in the end.Image Source: \cite{Gehr2018}}
    \label{fig:abstraction_DNN}
\end{figure}

Expanding over the Reluplex idea, the work in \cite{Gehr2018}, provide a more complete framework to prove formulas over a DNN. The key concept of $AI^2$ is the use of abstract interpretation to approximate mathematical functions with an infinity set of behaviors into logical functions witch are finite and consequently computable. To evaluate a DNN they propose to over-approximate the model in the abstract domain with the use of logical formulas capable of capturing certain shapes. \autoref{fig:abstraction_DNN} shows the exemplification of how the abstraction of the layers is used to evaluate properties in the DNN.

Building on the work of \cite{Gehr2018}, \textit{Singh et al.} \cite{Singh2018} published a work in which not only the ReLU activation function is available, but also Sigmoid and TanH. In their work, they implement a parallel version of the layer transformation which improved significantly the verification speed.

\subsubsection{Estimating the lower bound} 
In their work,\textit{Hein et al.} \cite{hein2017formal} the search of model robustness through the optics of formal guarantees. They've proposed in their work, a proven lower bound which establishes the minimal necessary perturbation to change a model's decision.  \textit{"We provide a guarantee that the classifier decision does not change in a certain ball around the considered instance"} \cite{hein2017formal}. Moreover, based on the proposed Cross-Lipschitz Regularazation method, they show the increase in the adversarial robustness of the models trained with such regularization. The generated bound is defined as:
\begin{align*}
    &\left \| \delta \right \|_p \leq \max_{\epsilon>0} \min \Psi\\
    &\Psi = \{\min_{j \neq c} \frac{f_c(x) - f_j(x)}{ \max_{y \in B_p(x,\epsilon)} \left \| \nabla f_c(y) - \nabla f_j(y) \right \|_q} \}
\end{align*}
It is known that an exact solution for the optimization problem which leads to the certification of DNN's is intractable for large networks. In \cite{weng2018evaluating}, they propose CLEVER (Cross Lipschitz Extreme Value for nEtwork Robustness). In their work, the lower bound is defined as the minimum $\epsilon$ necessary to be added in the input to change the decision of the model in an adversary setting.


Based on extreme value theory, the CLEVER metric is attack agnostic and is capable of estimating the lower bound for an attack to be effective in any model. But it restricts itself on providing a certification, it only provides an estimate of the lower bound.
Improving from CLEVER, \cite{weng2018towards} provides a certified lower bound for multi-layer perceptrons restricted to ReLU activation. In \cite{Zhang2018}, CROWN is proposed extending the exact certification to general activations. In \cite{boopathy2019cnn} the same research group proposed the CNN-Cert a framework to certify more general DNN.

In \textit{Zhang et al.} \cite{zhang2019towards}, a mix between inner bound propagation and linear relaxation is proposed. Linear relaxation of Neural Networks is one of the most popular methods to provide certified defenses and uses linear programming to provide linear relaxation, also known as the (convex adversarial polytope). Even though such methods generate an implementation that is tractable and solvable, they are still very demanding of computational resources. IBP on the contrary is less complex and brings more efficiency when optimizing verifiable networks.  Also known as interval bound propagation (IBP), are in general loose during the initial training phase which generates instabilities in the training and makes the model very sensitive to the hyper-parameters.

The model proposed by Zhang et. al. in \cite{zhang2019towards} unifies linear-relaxation and IBP. They generate a model which is very efficient for low output dimensions. It is used the convex relaxation in the backward pass of the bound, and the IBP in the forward bound pass of ther network. The optimization problem solved in CROWN-IBP can be defined as:
\begin{align*}
    &\min_\theta \mathbb{E}_{(x,y) \in \mathcal{D}} [\lambda \mathcal{L}(x,y,\theta)+(1-\lambda)\mathcal{L}(-(\Psi)+\Phi),y,\theta]\\
    &\Psi = (1-\beta)\underline{m}_{IBP}(x,\epsilon)\\
    &\Phi = \beta\underline{m}_{CROWN-IBP}(x,\epsilon)
\end{align*}
in which $\Psi$ is the IBP bound, $\Phi$ is the CROWN-IBP bound, and $\underline{m}(x,\epsilon)$ is the combination of both bounds.

\subsubsection{Upper Bounding the adversarial Loss} In the works of \cite{Wong2018} and \cite{Raghunathan2018} the certification of robustness is searched through means of defining an upper bound for the adversarial loss. For an adversarial loss defined as:
\begin{align*}
    \mathcal{L}_{adv} =& \max_{x'}\{max_{i\neq y} Z_i(x')-Z_y(x')\}\\ &\text{subj. to } x' \in B_\epsilon(x)
\end{align*}
both try to find a larger certificate $C(x,F)$ when compared to the loss of the perturbed example. If the certificate is smaller than 0, it is guaranteed that the true label will have the bigger score, and it can be stated that within this distance the model is safe. The works differ on the means to find the certificate. \cite{Wong2018} transforms the problem into a linear programming problem. \cite{Raghunathan2018} derives the certificate using semidefinite programming. An upper-bound estimation based on statistical methods is also proposed in \textit{Webb et al.} \cite{webb2018statistical}.




\subsubsection{Randomized Smoothing} Randomized Smoothing is a set of algorithms based on a mathematical formalist inspired in cryptography, \textit{differential privacy} (DP). This set of algorithms explore the connection between DP and robustness against norm-bounded adversarial examples in ML.

A classifier $f:R^d \rightarrow [0,1]^k$ which maps the input $x$ with probability $[0,1]$ to any of the $d$ classes, is said to be $\epsilon$-robust at $x$ if:
\begin{equation*}
    F(x+\delta) = c(x), \forall \delta:\mynorm{\delta}\leq \epsilon
\end{equation*}
moreover, if $f$ is $L-Lipschitz$ then $f$ is $\epsilon$-robust at $x$ with:
\begin{equation*}
    \epsilon = \frac{1}{2L}(P_A-P_B)
\end{equation*}
where, $P_A = \max_{i\in[k]} f_i(x)$ and $P_B$ is the second-max. Neural networks are known for being non-lipschitz. Recent work, from \textit{Lecuyer et al.} \cite{lecuyer2019certified} have proposed to smoothing the classifier, as in:
\begin{equation*}
    \hat{f}(x) = \mathbb{E}_{Z\sim\mathcal{N}(0,I_d)}(f(x+\sigma Z))
\end{equation*}
which is proven to be lipschitz.

\begin{figure}
    \centering
    \includegraphics[width=0.5\linewidth]{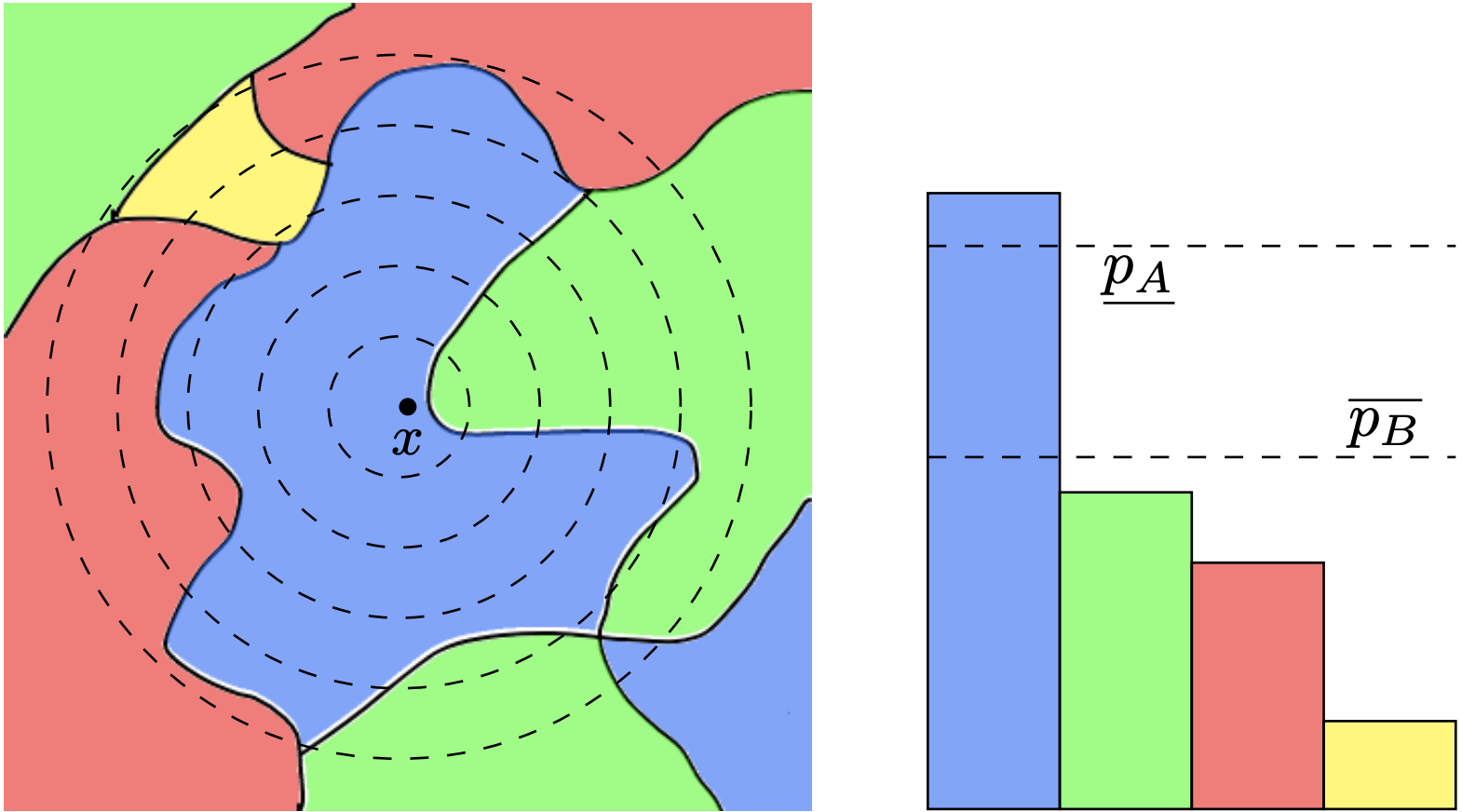}
    \caption{The smoothed classifier at an input x. In the left the decision boundaries for a classifier $f$, for each class represented in differente colours. The doted lines represents the sets of the distribution $\mathcal{N}(x,\sigma I)$. In the right the class probabilities, and the lower bound for class $p_A$ and the upper bound for each other class $p_B$. Image Source: \cite{cohen2019certified}}
    \label{fig:randomized_smoothing}
\end{figure}

\textit{Cohen et al.} \cite{cohen2019certified} propose the use of a smoothed classifier to generate a tighter certification bound. The proposed certification radius is defined by:
\begin{equation*}
    \epsilon = \frac{\sigma}{2}(\Phi^{-1}(\underline{P_A})-\Phi^{-1}(\Bar{P_B})),
\end{equation*}
where $\underline{P_A}$ the lower bound for the top class probability, $\Bar{P_B}$ is the upper bound probability for all other classes, and $\Phi^{-1}(.)$ is the inverse of the standard Gaussian cumulative distribution function.  In which $\epsilon$ is large when the noise level $\sigma$ is high, the probability of top class $A$ is high, and the probability of each other class is low. \autoref{fig:randomized_smoothing} shows the decision boundaries and the certification radius proposed by \textit{Cohen et al.} \cite{cohen2019certified}.

In \cite{zhai2020macer}, the authors propose an attack-free and scalable method to train robust deep neural networks. They mostly build upon Randomized Smoothing. The randomized smoothing classifier is defined as $g(x) = \mathbb{E}_\eta f(x+\eta)$, in which $\eta \sim \mathcal{N}(0,\sigma^2\text{\textbf{I}})$. Different from adversarial training as proposed in the original technique, the authors propose to robustly optimize including  certification radius as part of the defined objective.

\subsubsection{MMR-Universal} In \cite{croce2019provable}, a provable defense against all $l_p$-norms are proposed, for all $p\geq 1$. The case studied is the non-trivial case in which none of the $l_p$ balls are contained in the others. For an input with dimensions $d$, the robustness to an $l_p$ perturbation requires $d^{\frac{1}{p}-\frac{1}{q}}\epsilon_q > \epsilon_p > \epsilon_q$ for $p<q$. In which $\epsilon_p$ is the radius of the ball defined by the $l_p$-norm. In their paper a minimal $l_p$ norm of the complement of the union of $l_1$, $l_\infty$ norm, and its convex hull is derived as: 
\begin{equation*}
    \min_{x\in\mathbb{R}^d \text{\textbackslash} C} \left \| x \right \|_p = \frac{\epsilon_1}{(\frac{\epsilon_1}{\epsilon_\infty}-\alpha +\alpha^q)^{\frac{1}{q}}}
\end{equation*}
for $C$ the convex hull formed by the union of the $l_1$ and $l_\infty$ norm balls, $\alpha = \frac{\epsilon_1}{\epsilon_\infty} - \left \lfloor \frac{\epsilon_1}{\epsilon_\infty} \right \rfloor$, $\frac{1}{p}+\frac{1}{q} =1$, $d\geq2$ and $\epsilon_1 \in (\epsilon_\infty,d\epsilon_\infty)$. Based on this derivation, a lower bound for the robustness at point $x$ is defined and a regularizer term is expressed in term of the distances and the lower bound MMR-Unirversal. The regularizer is then incorporated in the loss function to improve the robustness of the model:
\begin{align*}
    \mathcal{L}(x_i,y_i) = &\frac{1}{T}\sum_{i=1}^T \mathcal{L}_{c-e}(f(x_i),y_i)\\&+MMR\text{-}Universal(x_i)
\end{align*}

During the optimization, the regularizer aims at pushing both the polytope boundaries and the decision hyperplanes farther than $l_\infty$ and $l_1$ distances from the training point $x$, to achieve robustness close or better than $l_\infty$ and $l_1$ respectively.

\section{Challenges and Future Opportunities}
\label{sec:ooportunities_challenges}

We've discussed and presented results in three methods for generating robust machine learning algorithms: adversarial (re)training, regularization, and certified defenses. The search for an optimal method to strengthen DL algorithms against adversaries has a solid structure but still requires a significant effort to achieve the major objective. While a great number of new algorithms have been published every year, both in the realm of attacks and defenses, no algorithm based on adversarial (re)Training or attack generation can claim to be the final and optimal method. At the speed new defenses arise, attackers exploit the gradient or other nuances from these defensive algorithms to generate their effective low norm perturbation. The reason for this arms race, is the fact that the defenses are not absolute, or that while trying to solve maximization in \autoref{eq:min_max_adversarial}, only an empirical approximation is used, and no global optimality is achieved. Moreover as the algorithms fail to account for all possible scenarios, there will always be an example in the neighborhood of $x$, such that $\mynorm{x-x'}_p\leq \delta$. 

Other than optimizing \autoref{eq:min_max_adversarial}, with an empirical solution, certified defense mechanisms present a formal alternative to achieve robust deep learning models, but this certified solution comes with the cost of efficiency, and high computational cost. It derives from \autoref{eq:layerwise_optimization}. The non-linear equality constraint for each layer of the neural network is unsolvable with standard techniques, and several methods have been proposed to achieve the optimal solution namely: linear relaxations, convex relaxations, interval propagation, abstract interpretation, mixed integer linear programming, or combination of these methods. We see great research opportunities and challenges in further improving such algorithms. They are in the early stages of development, in which the abstractions and formal approximations for the non-linearity constraints shown in \autoref{eq:layerwise_optimization} are not optimized for parallel computation as their numerical counterparts are. Such restriction makes it almost impractical to have such mechanisms to be applied in larger datasets. More in the topic, approximations of the optimization constraints either by a lower bound or upper bound has shown to speed-up the training process of such algorithms, but at the cost of having over-estimated bounds for the maximum allowed perturbation. These algorithms have not yet demonstrated high accuracy in large datasets corrupted by adversaries, or smaller datasets with high level of corruption. Part of the issue comes from the imposed convex relaxations. As they are not tight enough, it requires the verification algorithms to explore a bigger region, than actually necessary to verify the existence of adversarial examples in the neighborhood of the input.

Moreover as seen in \cite{ghiasi2020breaking}, even certified defenses can be broken when a big enough disturbance is applied to the model. It is arguable that even with the rigorous mathematical formulation of the defenses and certifications the constraint imposed by $l_p$ norm is weak. Most of the models can not achieve certifications beyond $\epsilon = 0.3$ disturbance in $l_2$ norm, while disturbances $\epsilon=4$ added to the target input are barely noticeable by human eyes, and $\epsilon = 100$, when applied to the original image are still easily classified by humans as belonging to the same class. As discussed by many authors, the perception of multi-dimensional space by human eyes goes beyond what the $l_p$ norm is capable of capturing and synthesizing. It is yet to be proposed more comprehensive metrics and algorithms capable of capturing the correlation between pixels of an image or input data which can better translate to optimization algorithms how humans distinguish features of an input image. Such a metric would allow the optimization algorithms to have better intuition on the subtle variations introduced by adversaries in the input data.

As we seek to apply machine learning algorithms in safe-critical applications, a model that works most of the time is not enough to guarantee safety of such implementations. It is imperative to know the operational restrictions of the algorithm, and the level of corruption they can safely handle. For such, formally certifying the algorithms is crucial, but increasing the neighborhood around the input that the certification can be guaranteed is fundamental for the practical application of current available techiniques.

\section{Conclusion}
Training safe and robust DNN algorithms is essential for their usage in safe-critical applications. This paper studies strategies to implement adversary robustly trained algorithms towards guaranteeing safety in machine learning models. We initially provide a taxonomy to classify adversarial attacks and defenses. Moreover, we provide a mathematical formulation for the Robust Optimization problem and divide it into a set of sub-categories which are: Adversarial (re)Training, Regularization Approach, and Certified Defenses. With the objective of elucidating methods to approximate the maximization problem, we present the most recent and important results in adversarial example generation. Furthermore, we describe several defense mechanisms that incorporate adversarial (re)Training as their main defense against perturbations or add regularization terms that change the behavior of the gradient, making it harder for attackers to achieve their objective. In addition, we surveyed methods which formally derive certificates of robustness by exactly solving the optimization problem or by approximations using upper or lower bounds. 
\label{sec:conclusion}

\bibliographystyle{IEEEtran}
\bibliography{references.bib}

\begin{thebibliography}{100}
\providecommand{\url}[1]{#1}
\csname url@samestyle\endcsname
\providecommand{\newblock}{\relax}
\providecommand{\bibinfo}[2]{#2}
\providecommand{\BIBentrySTDinterwordspacing}{\spaceskip=0pt\relax}
\providecommand{\BIBentryALTinterwordstretchfactor}{4}
\providecommand{\BIBentryALTinterwordspacing}{\spaceskip=\fontdimen2\font plus
\BIBentryALTinterwordstretchfactor\fontdimen3\font minus
  \fontdimen4\font\relax}
\providecommand{\BIBforeignlanguage}[2]{{%
\expandafter\ifx\csname l@#1\endcsname\relax
\typeout{** WARNING: IEEEtran.bst: No hyphenation pattern has been}%
\typeout{** loaded for the language `#1'. Using the pattern for}%
\typeout{** the default language instead.}%
\else
\language=\csname l@#1\endcsname
\fi
#2}}
\providecommand{\BIBdecl}{\relax}
\BIBdecl

\bibitem{lecun2015deep}
Y.~LeCun, Y.~Bengio, and G.~Hinton, ``Deep learning,'' \emph{nature}, vol. 521,
  no. 7553, pp. 436--444, 2015.

\bibitem{greenspan2016guest}
H.~Greenspan, B.~Van~Ginneken, and R.~M. Summers, ``Guest editorial deep
  learning in medical imaging: Overview and future promise of an exciting new
  technique,'' \emph{IEEE Transactions on Medical Imaging}, vol.~35, no.~5, pp.
  1153--1159, 2016.

\bibitem{stewart2015jetstream}
C.~A. Stewart, T.~M. Cockerill, I.~Foster, D.~Hancock, N.~Merchant,
  E.~Skidmore, D.~Stanzione, J.~Taylor, S.~Tuecke, G.~Turner \emph{et~al.},
  ``Jetstream: a self-provisioned, scalable science and engineering cloud
  environment,'' in \emph{Proceedings of the 2015 XSEDE Conference: Scientific
  Advancements Enabled by Enhanced Cyberinfrastructure}, 2015, pp. 1--8.

\bibitem{towns2014xsede}
J.~Towns, T.~Cockerill, M.~Dahan, I.~Foster, K.~Gaither, A.~Grimshaw,
  V.~Hazlewood, S.~Lathrop, D.~Lifka, G.~D. Peterson \emph{et~al.}, ``Xsede:
  accelerating scientific discovery,'' \emph{Computing in science \&
  engineering}, vol.~16, no.~5, pp. 62--74, 2014.

\bibitem{das2018distributed}
A.~Das, W.-M. Lin, and P.~Rad, ``A distributed secure machine-learning cloud
  architecture for semantic analysis,'' in \emph{Applied Cloud Deep Semantic
  Recognition}.\hskip 1em plus 0.5em minus 0.4em\relax Auerbach Publications,
  2018, pp. 145--174.

\bibitem{das2019distributed}
A.~Das, P.~Rad, K.-K.~R. Choo, B.~Nouhi, J.~Lish, and J.~Martel, ``Distributed
  machine learning cloud teleophthalmology iot for predicting amd disease
  progression,'' \emph{Future Generation Computer Systems}, vol.~93, pp.
  486--498, 2019.

\bibitem{yang2017biren}
B.~Yang, F.~Liu, C.~Ren, Z.~Ouyang, Z.~Xie, X.~Bo, and W.~Shu, ``Biren:
  predicting enhancers with a deep-learning-based model using the dna sequence
  alone,'' \emph{Bioinformatics}, vol.~33, no.~13, pp. 1930--1936, 2017.

\bibitem{gebru2017using}
T.~Gebru, J.~Krause, Y.~Wang, D.~Chen, J.~Deng, E.~L. Aiden, and L.~Fei-Fei,
  ``Using deep learning and google street view to estimate the demographic
  makeup of neighborhoods across the united states,'' \emph{Proceedings of the
  National Academy of Sciences}, vol. 114, no.~50, pp. 13\,108--13\,113, 2017.

\bibitem{gabbard2018matching}
H.~Gabbard, M.~Williams, F.~Hayes, and C.~Messenger, ``Matching matched
  filtering with deep networks for gravitational-wave astronomy,''
  \emph{Physical review letters}, vol. 120, no.~14, p. 141103, 2018.

\bibitem{ebadi2019implicit}
N.~Ebadi, B.~Lwowski, M.~Jaloli, and P.~Rad, ``Implicit life event discovery
  from call transcripts using temporal input transformation network,''
  \emph{IEEE Access}, vol.~7, pp. 172\,178--172\,189, 2019.

\bibitem{amodei2016deep}
D.~Amodei, S.~Ananthanarayanan, R.~Anubhai, J.~Bai, E.~Battenberg, C.~Case,
  J.~Casper, B.~Catanzaro, Q.~Cheng, G.~Chen \emph{et~al.}, ``Deep speech 2:
  End-to-end speech recognition in english and mandarin,'' in
  \emph{International conference on machine learning}, 2016, pp. 173--182.

\bibitem{voulodimos2018deep}
A.~Voulodimos, N.~Doulamis, A.~Doulamis, and E.~Protopapadakis, ``Deep learning
  for computer vision: A brief review,'' \emph{Computational intelligence and
  neuroscience}, vol. 2018, 2018.

\bibitem{redmon2016you}
J.~Redmon, S.~Divvala, R.~Girshick, and A.~Farhadi, ``You only look once:
  Unified, real-time object detection,'' in \emph{Proceedings of the IEEE
  conference on computer vision and pattern recognition}, 2016, pp. 779--788.

\bibitem{dai2016r}
J.~Dai, Y.~Li, K.~He, and J.~Sun, ``R-fcn: Object detection via region-based
  fully convolutional networks,'' in \emph{Advances in neural information
  processing systems}, 2016, pp. 379--387.

\bibitem{varol2017long}
G.~Varol, I.~Laptev, and C.~Schmid, ``Long-term temporal convolutions for
  action recognition,'' \emph{IEEE transactions on pattern analysis and machine
  intelligence}, vol.~40, no.~6, pp. 1510--1517, 2017.

\bibitem{9043480}
N.~Bendre, N.~Ebadi, J.~J. Prevost, and P.~Najafirad, ``Human action
  performance using deep neuro-fuzzy recurrent attention model,'' \emph{IEEE
  Access}, vol.~8, pp. 57\,749--57\,761, 2020.

\bibitem{cao2017realtime}
Z.~Cao, T.~Simon, S.-E. Wei, and Y.~Sheikh, ``Realtime multi-person 2d pose
  estimation using part affinity fields,'' in \emph{Proceedings of the IEEE
  Conference on Computer Vision and Pattern Recognition}, 2017, pp. 7291--7299.

\bibitem{newell2016stacked}
A.~Newell, K.~Yang, and J.~Deng, ``Stacked hourglass networks for human pose
  estimation,'' in \emph{European conference on computer vision}.\hskip 1em
  plus 0.5em minus 0.4em\relax Springer, 2016, pp. 483--499.

\bibitem{badrinarayanan2017segnet}
V.~Badrinarayanan, A.~Kendall, and R.~Cipolla, ``Segnet: A deep convolutional
  encoder-decoder architecture for image segmentation,'' \emph{IEEE
  transactions on pattern analysis and machine intelligence}, vol.~39, no.~12,
  pp. 2481--2495, 2017.

\bibitem{chen2017deeplab}
L.-C. Chen, G.~Papandreou, I.~Kokkinos, K.~Murphy, and A.~L. Yuille, ``Deeplab:
  Semantic image segmentation with deep convolutional nets, atrous convolution,
  and fully connected crfs,'' \emph{IEEE transactions on pattern analysis and
  machine intelligence}, vol.~40, no.~4, pp. 834--848, 2017.

\bibitem{held2016learning}
D.~Held, S.~Thrun, and S.~Savarese, ``Learning to track at 100 fps with deep
  regression networks,'' in \emph{European Conference on Computer
  Vision}.\hskip 1em plus 0.5em minus 0.4em\relax Springer, 2016, pp. 749--765.

\bibitem{krizhevsky2012imagenet}
A.~Krizhevsky, I.~Sutskever, and G.~E. Hinton, ``Imagenet classification with
  deep convolutional neural networks,'' in \emph{Advances in neural information
  processing systems}, 2012, pp. 1097--1105.

\bibitem{litjens2017survey}
G.~Litjens, T.~Kooi, B.~E. Bejnordi, A.~A.~A. Setio, F.~Ciompi, M.~Ghafoorian,
  J.~A. Van Der~Laak, B.~Van~Ginneken, and C.~I. S{\'a}nchez, ``A survey on
  deep learning in medical image analysis,'' \emph{Medical image analysis},
  vol.~42, pp. 60--88, 2017.

\bibitem{abadi2016tensorflow}
M.~Abadi, P.~Barham, J.~Chen, Z.~Chen, A.~Davis, J.~Dean, M.~Devin,
  S.~Ghemawat, G.~Irving, M.~Isard \emph{et~al.}, ``Tensorflow: A system for
  large-scale machine learning,'' in \emph{12th $\{$USENIX$\}$ Symposium on
  Operating Systems Design and Implementation ($\{$OSDI$\}$ 16)}, 2016, pp.
  265--283.

\bibitem{paszke2019pytorch}
A.~Paszke, S.~Gross, F.~Massa, A.~Lerer, J.~Bradbury, G.~Chanan, T.~Killeen,
  Z.~Lin, N.~Gimelshein, L.~Antiga \emph{et~al.}, ``Pytorch: An imperative
  style, high-performance deep learning library,'' in \emph{Advances in Neural
  Information Processing Systems}, 2019, pp. 8024--8035.

\bibitem{keahey2019chameleon}
K.~Keahey, P.~Riteau, D.~Stanzione, T.~Cockerill, J.~Mambretti, P.~Rad, and
  P.~Ruth, ``Chameleon: a scalable production testbed for computer science
  research,'' in \emph{Contemporary High Performance Computing}.\hskip 1em plus
  0.5em minus 0.4em\relax CRC Press, 2019, pp. 123--148.

\bibitem{he2016deep}
K.~He, X.~Zhang, S.~Ren, and J.~Sun, ``Deep residual learning for image
  recognition,'' in \emph{Proceedings of the IEEE conference on computer vision
  and pattern recognition}, 2016, pp. 770--778.

\bibitem{szegedy2016rethinking}
C.~Szegedy, V.~Vanhoucke, S.~Ioffe, J.~Shlens, and Z.~Wojna, ``Rethinking the
  inception architecture for computer vision,'' in \emph{Proceedings of the
  IEEE conference on computer vision and pattern recognition}, 2016, pp.
  2818--2826.

\bibitem{szegedy2017inception}
C.~Szegedy, S.~Ioffe, V.~Vanhoucke, and A.~A. Alemi, ``Inception-v4,
  inception-resnet and the impact of residual connections on learning,'' in
  \emph{Thirty-first AAAI conference on artificial intelligence}, 2017.

\bibitem{akhtar2018threat}
N.~Akhtar and A.~Mian, ``Threat of adversarial attacks on deep learning in
  computer vision: A survey,'' \emph{IEEE Access}, vol.~6, pp.
  14\,410--14\,430, 2018.

\bibitem{bojarski2016end}
M.~Bojarski, D.~Del~Testa, D.~Dworakowski, B.~Firner, B.~Flepp, P.~Goyal, L.~D.
  Jackel, M.~Monfort, U.~Muller, J.~Zhang \emph{et~al.}, ``End to end learning
  for self-driving cars,'' \emph{arXiv preprint arXiv:1604.07316}, 2016.

\bibitem{silva2017multi}
S.~Silva, R.~Suresh, F.~Tao, J.~Votion, and Y.~Cao, ``A multi-layer k-means
  approach for multi-sensor data pattern recognition in multi-target
  localization,'' \emph{arXiv preprint arXiv:1705.10757}, 2017.

\bibitem{silva2019cooperative}
S.~H. Silva, P.~Rad, N.~Beebe, K.-K.~R. Choo, and M.~Umapathy, ``Cooperative
  unmanned aerial vehicles with privacy preserving deep vision for real-time
  object identification and tracking,'' \emph{Journal of Parallel and
  Distributed Computing}, vol. 131, pp. 147--160, 2019.

\bibitem{gu2017deep}
S.~Gu, E.~Holly, T.~Lillicrap, and S.~Levine, ``Deep reinforcement learning for
  robotic manipulation with asynchronous off-policy updates,'' in \emph{2017
  IEEE international conference on robotics and automation (ICRA)}.\hskip 1em
  plus 0.5em minus 0.4em\relax IEEE, 2017, pp. 3389--3396.

\bibitem{silva2020temporal}
S.~H. {Silva}, A.~{Alaeddini}, and P.~{Najafirad}, ``Temporal graph traversals
  using reinforcement learning with proximal policy optimization,'' \emph{IEEE
  Access}, vol.~8, pp. 63\,910--63\,922, 2020.

\bibitem{lansley2019seen}
M.~Lansley, N.~Polatidis, S.~Kapetanakis, K.~Amin, G.~Samakovitis, and
  M.~Petridis, ``Seen the villains: Detecting social engineering attacks using
  case-based reasoning and deep learning,'' in \emph{Workshops Proceedings for
  the Twenty-seventh International Conference on Case-Based Reasoning:
  Case-based reasoning and deep learning workshop}, 2019.

\bibitem{kwon2017survey}
D.~Kwon, H.~Kim, J.~Kim, S.~C. Suh, I.~Kim, and K.~J. Kim, ``A survey of deep
  learning-based network anomaly detection,'' \emph{Cluster Computing}, pp.
  1--13, 2017.

\bibitem{de2019implementation}
G.~De~La~Torre, P.~Rad, and K.-K.~R. Choo, ``Implementation of deep packet
  inspection in smart grids and industrial internet of things: Challenges and
  opportunities,'' \emph{Journal of Network and Computer Applications}, 2019.

\bibitem{kepuska2018next}
V.~Kepuska and G.~Bohouta, ``Next-generation of virtual personal assistants
  (microsoft cortana, apple siri, amazon alexa and google home),'' in
  \emph{2018 IEEE 8th Annual Computing and Communication Workshop and
  Conference (CCWC)}.\hskip 1em plus 0.5em minus 0.4em\relax IEEE, 2018, pp.
  99--103.

\bibitem{cheng2016wide}
H.-T. Cheng, L.~Koc, J.~Harmsen, T.~Shaked, T.~Chandra, H.~Aradhye,
  G.~Anderson, G.~Corrado, W.~Chai, M.~Ispir \emph{et~al.}, ``Wide \& deep
  learning for recommender systems,'' in \emph{Proceedings of the 1st workshop
  on deep learning for recommender systems}, 2016, pp. 7--10.

\bibitem{de2018driverless}
G.~De~La~Torre, P.~Rad, and K.-K.~R. Choo, ``Driverless vehicle security:
  Challenges and future research opportunities,'' \emph{Future Generation
  Computer Systems}, 2018.

\bibitem{mnih2015human}
V.~Mnih, K.~Kavukcuoglu, D.~Silver, A.~A. Rusu, J.~Veness, M.~G. Bellemare,
  A.~Graves, M.~Riedmiller, A.~K. Fidjeland, G.~Ostrovski \emph{et~al.},
  ``Human-level control through deep reinforcement learning,'' \emph{Nature},
  vol. 518, no. 7540, pp. 529--533, 2015.

\bibitem{szegedy2013intriguing}
C.~Szegedy, W.~Zaremba, I.~Sutskever, J.~B. Estrach, D.~Erhan, I.~Goodfellow,
  and R.~Fergus, ``Intriguing properties of neural networks,'' in \emph{2nd
  International Conference on Learning Representations, ICLR 2014}, 2014.

\bibitem{nguyen2015deep}
A.~Nguyen, J.~Yosinski, and J.~Clune, ``Deep neural networks are easily fooled:
  High confidence predictions for unrecognizable images,'' in \emph{Proceedings
  of the IEEE conference on computer vision and pattern recognition}, 2015, pp.
  427--436.

\bibitem{biggio2014security}
B.~Biggio, I.~Corona, B.~Nelson, B.~I. Rubinstein, D.~Maiorca, G.~Fumera,
  G.~Giacinto, and F.~Roli, ``Security evaluation of support vector machines in
  adversarial environments,'' in \emph{Support Vector Machines
  Applications}.\hskip 1em plus 0.5em minus 0.4em\relax Springer, 2014, pp.
  105--153.

\bibitem{chakraborty2018adversarial}
A.~Chakraborty, M.~Alam, V.~Dey, A.~Chattopadhyay, and D.~Mukhopadhyay,
  ``Adversarial attacks and defences: A survey,'' \emph{arXiv preprint
  arXiv:1810.00069}, 2018.

\bibitem{chacon2019deep}
H.~Chacon, S.~Silva, and P.~Rad, ``Deep learning poison data attack
  detection,'' in \emph{2019 IEEE 31st International Conference on Tools with
  Artificial Intelligence (ICTAI)}.\hskip 1em plus 0.5em minus 0.4em\relax
  IEEE, 2019, pp. 971--978.

\bibitem{das2020opportunities}
A.~Das and P.~Rad, ``Opportunities and challenges in explainable artificial
  intelligence (xai): A survey,'' \emph{arXiv preprint arXiv:2006.11371}, 2020.

\bibitem{carlini2017towards}
N.~Carlini and D.~Wagner, ``Towards evaluating the robustness of neural
  networks,'' in \emph{2017 Ieee symposium on security and privacy (sp)}.\hskip
  1em plus 0.5em minus 0.4em\relax IEEE, 2017, pp. 39--57.

\bibitem{yuan2019adversarial}
X.~Yuan, P.~He, Q.~Zhu, and X.~Li, ``Adversarial examples: Attacks and defenses
  for deep learning,'' \emph{IEEE transactions on neural networks and learning
  systems}, vol.~30, no.~9, pp. 2805--2824, 2019.

\bibitem{vorobeychik2018adversarial}
Y.~Vorobeychik and M.~Kantarcioglu, \emph{Adversarial machine learning}.\hskip
  1em plus 0.5em minus 0.4em\relax Morgan \& Claypool Publishers, 2018.

\bibitem{chen2017zoo}
P.-Y. Chen, H.~Zhang, Y.~Sharma, J.~Yi, and C.-J. Hsieh, ``Zoo: Zeroth order
  optimization based black-box attacks to deep neural networks without training
  substitute models,'' in \emph{Proceedings of the 10th ACM Workshop on
  Artificial Intelligence and Security}, 2017, pp. 15--26.

\bibitem{xu2019adversarial}
H.~Xu, Y.~Ma, H.-C. Liu, D.~Deb, H.~Liu, J.-L. Tang, and A.~K. Jain,
  ``Adversarial attacks and defenses in images, graphs and text: A review,''
  \emph{International Journal of Automation and Computing}, vol.~17, no.~2, pp.
  151--178, 2020.

\bibitem{Madry2018}
A.~Madry, A.~Makelov, L.~Schmidt, D.~Tsipras, and A.~Vladu, ``Towards deep
  learning models resistant to adversarial attacks,'' in \emph{International
  Conference on Learning Representations}, 2018.

\bibitem{Xie2019}
C.~Xie, Y.~Wu, L.~v.~d. Maaten, A.~L. Yuille, and K.~He, ``Feature denoising
  for improving adversarial robustness,'' in \emph{Proceedings of the IEEE
  Conference on Computer Vision and Pattern Recognition}, 2019, pp. 501--509.

\bibitem{Goodfellow2015}
I.~J. Goodfellow, J.~Shlens, and C.~Szegedy, ``{Explaining and harnessing
  adversarial examples},'' in \emph{3rd International Conference on Learning
  Representations, ICLR 2015 - Conference Track Proceedings}, 2015.

\bibitem{Moosavi-Dezfooli2016}
S.-M. Moosavi-Dezfooli, A.~Fawzi, and P.~Frossard, ``Deepfool: a simple and
  accurate method to fool deep neural networks,'' in \emph{Proceedings of the
  IEEE conference on computer vision and pattern recognition}, 2016, pp.
  2574--2582.

\bibitem{Papernot2016}
N.~Papernot, P.~McDaniel, S.~Jha, M.~Fredrikson, Z.~B. Celik, and A.~Swami,
  ``The limitations of deep learning in adversarial settings,'' in \emph{2016
  IEEE European symposium on security and privacy (EuroS\&P)}.\hskip 1em plus
  0.5em minus 0.4em\relax IEEE, 2016, pp. 372--387.

\bibitem{Kurakin2017}
A.~Kurakin, I.~J. Goodfellow, and S.~Bengio, ``{Adversarial machine learning at
  scale},'' in \emph{5th International Conference on Learning Representations,
  ICLR 2017 - Conference Track Proceedings}, 2017.

\bibitem{Carlini2017}
N.~Carlini and D.~Wagner, ``Towards evaluating the robustness of neural
  networks,'' in \emph{2017 ieee symposium on security and privacy (sp)}.\hskip
  1em plus 0.5em minus 0.4em\relax IEEE, 2017, pp. 39--57.

\bibitem{papernot2016distillation}
N.~Papernot, P.~McDaniel, X.~Wu, S.~Jha, and A.~Swami, ``Distillation as a
  defense to adversarial perturbations against deep neural networks,'' in
  \emph{2016 IEEE Symposium on Security and Privacy (SP)}.\hskip 1em plus 0.5em
  minus 0.4em\relax IEEE, 2016, pp. 582--597.

\bibitem{Carlini2018}
N.~Carlini, G.~Katz, C.~Barrett, and D.~L. Dill, ``{Ground-Truth Adversarial
  Examples},'' \emph{Iclr 2018}, 2018.

\bibitem{Katz2017}
G.~Katz, C.~Barrett, D.~L. Dill, K.~Julian, and M.~J. Kochenderfer, ``Reluplex:
  An efficient smt solver for verifying deep neural networks,'' in
  \emph{International Conference on Computer Aided Verification}.\hskip 1em
  plus 0.5em minus 0.4em\relax Springer, 2017, pp. 97--117.

\bibitem{Tjeng2019}
V.~Tjeng, K.~Y. Xiao, and R.~Tedrake, ``Evaluating robustness of neural
  networks with mixed integer programming,'' in \emph{International Conference
  on Learning Representations}, 2018.

\bibitem{Xiao2019}
K.~Y. Xiao, V.~Tjeng, N.~M.~M. Shafiullah, and A.~Madry, ``Training for faster
  adversarial robustness verification via inducing relu stability,'' in
  \emph{International Conference on Learning Representations}, 2018.

\bibitem{moosavi2017universal}
S.-M. Moosavi-Dezfooli, A.~Fawzi, O.~Fawzi, and P.~Frossard, ``Universal
  adversarial perturbations,'' in \emph{Proceedings of the IEEE conference on
  computer vision and pattern recognition}, 2017, pp. 1765--1773.

\bibitem{ghiasi2020breaking}
A.~Ghiasi, A.~Shafahi, and T.~Goldstein, ``Breaking certified defenses:
  Semantic adversarial examples with spoofed robustness certificates,'' in
  \emph{International Conference on Learning Representations}, 2019.

\bibitem{chen2018ead}
P.-Y. Chen, Y.~Sharma, H.~Zhang, J.~Yi, and C.-J. Hsieh, ``Ead: elastic-net
  attacks to deep neural networks via adversarial examples,'' in
  \emph{Thirty-second AAAI conference on artificial intelligence}, 2018.

\bibitem{jang2017objective}
U.~Jang, X.~Wu, and S.~Jha, ``Objective metrics and gradient descent algorithms
  for adversarial examples in machine learning,'' in \emph{Proceedings of the
  33rd Annual Computer Security Applications Conference}, 2017, pp. 262--277.

\bibitem{xiao2018spatially}
C.~Xiao, J.~Y. Zhu, B.~Li, W.~He, M.~Liu, and D.~Song, ``Spatially transformed
  adversarial examples,'' in \emph{6th International Conference on Learning
  Representations, ICLR 2018}, 2018.

\bibitem{engstrom2019exploring}
L.~Engstrom, B.~Tran, D.~Tsipras, L.~Schmidt, and A.~Madry, ``Exploring the
  landscape of spatial robustness,'' in \emph{ICML}, 2019.

\bibitem{odena2017conditional}
A.~Odena, C.~Olah, and J.~Shlens, ``Conditional image synthesis with auxiliary
  classifier gans,'' in \emph{Proceedings of the 34th International Conference
  on Machine Learning-Volume 70}.\hskip 1em plus 0.5em minus 0.4em\relax JMLR.
  org, 2017, pp. 2642--2651.

\bibitem{song2018constructing}
Y.~Song, R.~Shu, N.~Kushman, and S.~Ermon, ``Constructing unrestricted
  adversarial examples with generative models,'' in \emph{Advances in Neural
  Information Processing Systems}, 2018, pp. 8312--8323.

\bibitem{papernot2017practical}
N.~Papernot, P.~McDaniel, I.~Goodfellow, S.~Jha, Z.~B. Celik, and A.~Swami,
  ``Practical black-box attacks against machine learning,'' in
  \emph{Proceedings of the 2017 ACM on Asia conference on computer and
  communications security}, 2017, pp. 506--519.

\bibitem{su2019one}
J.~Su, D.~V. Vargas, and K.~Sakurai, ``One pixel attack for fooling deep neural
  networks,'' \emph{IEEE Transactions on Evolutionary Computation}, vol.~23,
  no.~5, pp. 828--841, 2019.

\bibitem{ilyas2017query}
A.~Ilyas, L.~Engstrom, A.~Athalye, and J.~Lin, ``Query-efficient black-box
  adversarial examples,'' \emph{arXiv preprint arXiv:1712.07113}, 2017.

\bibitem{chen2019hopskipjumpattack}
J.~Chen, M.~Jordan, and M.~Wainwright, ``Hopskipjumpattack: A query-efficient
  decision-based attack,'' in \emph{2020 IEEE Symposium on Security and Privacy
  (SP)}, pp. 668--685.

\bibitem{ru2020bayesopt}
B.~Ru, A.~Cobb, A.~Blaas, and Y.~Gal, ``Bayesopt adversarial attack,'' in
  \emph{International Conference on Learning Representations}, 2020.

\bibitem{gleave2019adversarial}
A.~Gleave, M.~Dennis, C.~Wild, N.~Kant, S.~Levine, and S.~Russell,
  ``Adversarial policies: Attacking deep reinforcement learning,'' in
  \emph{International Conference on Learning Representations}, 2019.

\bibitem{evtimov2017robust}
K.~{Eykholt}, I.~{Evtimov}, E.~{Fernandes}, B.~{Li}, A.~{Rahmati}, C.~{Xiao},
  A.~{Prakash}, T.~{Kohno}, and D.~{Song}, ``Robust physical-world attacks on
  deep learning visual classification,'' in \emph{2018 IEEE/CVF Conference on
  Computer Vision and Pattern Recognition}, 2018, pp. 1625--1634.

\bibitem{goswami2018unravelling}
G.~Goswami, N.~Ratha, A.~Agarwal, R.~Singh, and M.~Vatsa, ``Unravelling
  robustness of deep learning based face recognition against adversarial
  attacks,'' in \emph{Thirty-Second AAAI Conference on Artificial
  Intelligence}, 2018.

\bibitem{bose2018adversarial}
A.~J. Bose and P.~Aarabi, ``Adversarial attacks on face detectors using neural
  net based constrained optimization,'' in \emph{2018 IEEE 20th International
  Workshop on Multimedia Signal Processing (MMSP)}.\hskip 1em plus 0.5em minus
  0.4em\relax IEEE, 2018, pp. 1--6.

\bibitem{zhu2019generating}
Z.-A. Zhu, Y.-Z. Lu, and C.-K. Chiang, ``Generating adversarial examples by
  makeup attacks on face recognition,'' in \emph{2019 IEEE International
  Conference on Image Processing (ICIP)}.\hskip 1em plus 0.5em minus
  0.4em\relax IEEE, 2019, pp. 2516--2520.

\bibitem{dong2019efficient}
Y.~Dong, H.~Su, B.~Wu, Z.~Li, W.~Liu, T.~Zhang, and J.~Zhu, ``Efficient
  decision-based black-box adversarial attacks on face recognition,'' in
  \emph{Proceedings of the IEEE Conference on Computer Vision and Pattern
  Recognition}, 2019, pp. 7714--7722.

\bibitem{suciu2019exploring}
O.~Suciu, S.~E. Coull, and J.~Johns, ``Exploring adversarial examples in
  malware detection,'' in \emph{2019 IEEE Security and Privacy Workshops
  (SPW)}.\hskip 1em plus 0.5em minus 0.4em\relax IEEE, 2019, pp. 8--14.

\bibitem{chernikova2019adversarial}
A.~Chernikova and A.~Oprea, ``Adversarial examples for deep learning cyber
  security analytics,'' \emph{arXiv preprint arXiv:1909.10480}, 2019.

\bibitem{melis2017deep}
M.~Melis, A.~Demontis, B.~Biggio, G.~Brown, G.~Fumera, and F.~Roli, ``Is deep
  learning safe for robot vision? adversarial examples against the icub
  humanoid,'' in \emph{Proceedings of the IEEE International Conference on
  Computer Vision Workshops}, 2017, pp. 751--759.

\bibitem{biggio2013evasion}
B.~Biggio, I.~Corona, D.~Maiorca, B.~Nelson, N.~{\v{S}}rndi{\'c}, P.~Laskov,
  G.~Giacinto, and F.~Roli, ``Evasion attacks against machine learning at test
  time,'' in \emph{Joint European conference on machine learning and knowledge
  discovery in databases}.\hskip 1em plus 0.5em minus 0.4em\relax Springer,
  2013, pp. 387--402.

\bibitem{sitawarin2018darts}
C.~Sitawarin, A.~N. Bhagoji, A.~Mosenia, M.~Chiang, and P.~Mittal, ``Darts:
  Deceiving autonomous cars with toxic signs,'' \emph{arXiv preprint
  arXiv:1802.06430}, 2018.

\bibitem{morgulis2019fooling}
N.~Morgulis, A.~Kreines, S.~Mendelowitz, and Y.~Weisglass, ``Fooling a real car
  with adversarial traffic signs,'' \emph{arXiv preprint arXiv:1907.00374},
  2019.

\bibitem{cao2019adversarial}
Y.~Cao, C.~Xiao, B.~Cyr, Y.~Zhou, W.~Park, S.~Rampazzi, Q.~A. Chen, K.~Fu, and
  Z.~M. Mao, ``Adversarial sensor attack on lidar-based perception in
  autonomous driving,'' in \emph{Proceedings of the 2019 ACM SIGSAC Conference
  on Computer and Communications Security}, 2019, pp. 2267--2281.

\bibitem{jia2019fooling}
Y.~Jia, Y.~Lu, J.~Shen, Q.~A. Chen, H.~Chen, Z.~Zhong, and T.~Wei, ``Fooling
  detection alone is not enough: Adversarial attack against multiple object
  tracking,'' in \emph{International Conference on Learning Representations},
  2019.

\bibitem{brendel2017decision}
W.~Brendel, J.~Rauber, and M.~Bethge, ``Decision-based adversarial attacks:
  Reliable attacks against black-box machine learning models,'' in
  \emph{International Conference on Learning Representations}, 2018.

\bibitem{sarkar2017upset}
S.~Sarkar, A.~Bansal, U.~Mahbub, and R.~Chellappa, ``Upset and angri: Breaking
  high performance image classifiers,'' \emph{arXiv preprint arXiv:1707.01159},
  2017.

\bibitem{wang2020improving}
Y.~Wang, D.~Zou, J.~Yi, J.~Bailey, X.~Ma, and Q.~Gu, ``Improving adversarial
  robustness requires revisiting misclassified examples,'' in
  \emph{International Conference on Learning Representations}, 2020.

\bibitem{song2019robust}
C.~Song, K.~He, J.~Lin, L.~Wang, and J.~E. Hopcroft, ``Robust local features
  for improving the generalization of adversarial training,'' in
  \emph{International Conference on Learning Representations}, 2019.

\bibitem{wu2019defending}
T.~Wu, L.~Tong, and Y.~Vorobeychik, ``Defending against physically realizable
  attacks on image classification,'' in \emph{International Conference on
  Learning Representations}, 2020.

\bibitem{wong2020fast}
E.~Wong, L.~Rice, and J.~Z. Kolter, ``Fast is better than free: Revisiting
  adversarial training,'' in \emph{International Conference on Learning
  Representations}, 2019.

\bibitem{hu2020triple}
T.-K. Hu, T.~Chen, H.~Wang, and Z.~Wang, ``Triple wins: Boosting accuracy,
  robustness and efficiency together by enabling input-adaptive inference,'' in
  \emph{International Conference on Learning Representations}, 2019.

\bibitem{Chen2019}
H.~Chen, H.~Zhang, D.~Boning, and C.-J. Hsieh, ``Robust decision trees against
  adversarial examples,'' in \emph{International Conference on Machine
  Learning}, 2019, pp. 1122--1131.

\bibitem{Yang2019}
Y.~Yang, G.~Zhang, D.~Katabi, and Z.~Xu, ``Me-net: Towards effective
  adversarial robustness with matrix estimation,'' in \emph{International
  Conference on Machine Learning}, 2019, pp. 7025--7034.

\bibitem{Kannan2018}
H.~Kannan, A.~Kurakin, and I.~Goodfellow, ``Adversarial logit pairing,''
  \emph{arXiv preprint arXiv:1803.06373}, 2018.

\bibitem{Matyasko2018}
A.~Matyasko and L.-P. Chau, ``Improved network robustness with adversary
  critic,'' in \emph{Advances in Neural Information Processing Systems}, 2018,
  pp. 10\,578--10\,587.

\bibitem{Sinha2018}
A.~Sinha, H.~Namkoong, and J.~Duchi, ``Certifying some distributional
  robustness with principled adversarial training,'' in \emph{International
  Conference on Learning Representations}, 2018.

\bibitem{sen2020empir}
S.~Sen, B.~Ravindran, and A.~Raghunathan, ``Empir: Ensembles of mixed precision
  deep networks for increased robustness against adversarial attacks,'' in
  \emph{International Conference on Learning Representations}, 2019.

\bibitem{Liu2018}
X.~Liu, M.~Cheng, H.~Zhang, and C.-J. Hsieh, ``Towards robust neural networks
  via random self-ensemble,'' in \emph{Proceedings of the European Conference
  on Computer Vision (ECCV)}, 2018, pp. 369--385.

\bibitem{Tramer2018}
F.~Tram{\`e}r, A.~Kurakin, N.~Papernot, I.~Goodfellow, D.~Boneh, and
  P.~McDaniel, ``Ensemble adversarial training: Attacks and defenses,'' in
  \emph{International Conference on Learning Representations}, 2018.

\bibitem{Wang2019}
B.~Wang, S.~Webb, and T.~Rainforth, ``Statistically robust neural network
  classification,'' \emph{arXiv preprint arXiv:1912.04884}, 2019.

\bibitem{zhai2020macer}
R.~Zhai, C.~Dan, D.~He, H.~Zhang, B.~Gong, P.~Ravikumar, C.-J. Hsieh, and
  L.~Wang, ``Macer: Attack-free and scalable robust training via maximizing
  certified radius,'' in \emph{International Conference on Learning
  Representations}, 2019.

\bibitem{croce2019provable}
F.~Croce and M.~Hein, ``Provable robustness against all adversarial $ l\_p
  $-perturbations for $p$ $geq 1$,'' in \emph{International Conference on
  Learning Representations}, 2019.

\bibitem{Raghunathan2018}
A.~Raghunathan, J.~Steinhardt, and P.~Liang, ``Certified defenses against
  adversarial examples,'' in \emph{International Conference on Learning
  Representations}, 2018.

\bibitem{Wong2018}
E.~Wong and Z.~Kolter, ``Provable defenses against adversarial examples via the
  convex outer adversarial polytope,'' in \emph{International Conference on
  Machine Learning}, 2018, pp. 5286--5295.

\bibitem{boopathy2019cnn}
A.~Boopathy, T.-W. Weng, P.-Y. Chen, S.~Liu, and L.~Daniel, ``Cnn-cert: An
  efficient framework for certifying robustness of convolutional neural
  networks,'' in \emph{Proceedings of the AAAI Conference on Artificial
  Intelligence}, vol.~33, 2019, pp. 3240--3247.

\bibitem{Zhang2018}
H.~Zhang, T.-W. Weng, P.-Y. Chen, C.-J. Hsieh, and L.~Daniel, ``Efficient
  neural network robustness certification with general activation functions,''
  in \emph{Advances in neural information processing systems}, 2018, pp.
  4939--4948.

\bibitem{weng2018towards}
T.-W. Weng, H.~Zhang, H.~Chen, Z.~Song, C.-J. Hsieh, L.~Daniel, and I.~Dhillon,
  ``Towards fast computation of certified robustness for relu networks,'' in
  \emph{International Conference on Machine Learning (ICML)}, 2018.

\bibitem{weng2018evaluating}
T.-W. Weng, H.~Zhang, P.-Y. Chen, J.~Yi, D.~Su, Y.~Gao, C.-J. Hsieh, and
  L.~Daniel, ``Evaluating the robustness of neural networks: An extreme value
  theory approach,'' \emph{arXiv preprint arXiv:1801.10578}, 2018.

\bibitem{hein2017formal}
M.~Hein and M.~Andriushchenko, ``Formal guarantees on the robustness of a
  classifier against adversarial manipulation,'' in \emph{Advances in Neural
  Information Processing Systems}, 2017, pp. 2266--2276.

\bibitem{Singh2018}
G.~Singh, T.~Gehr, M.~Mirman, M.~P{\"u}schel, and M.~Vechev, ``Fast and
  effective robustness certification,'' in \emph{Advances in Neural Information
  Processing Systems}, 2018, pp. 10\,802--10\,813.

\bibitem{Gehr2018}
T.~Gehr, M.~Mirman, D.~Drachsler-Cohen, P.~Tsankov, S.~Chaudhuri, and
  M.~Vechev, ``Ai2: Safety and robustness certification of neural networks with
  abstract interpretation,'' in \emph{2018 IEEE Symposium on Security and
  Privacy (SP)}.\hskip 1em plus 0.5em minus 0.4em\relax IEEE, 2018, pp. 3--18.

\bibitem{mao2019metric}
C.~Mao, Z.~Zhong, J.~Yang, C.~Vondrick, and B.~Ray, ``Metric learning for
  adversarial robustness,'' in \emph{Advances in Neural Information Processing
  Systems}, 2019, pp. 478--489.

\bibitem{Zhang2019}
H.~Zhang, Y.~Yu, J.~Jiao, E.~Xing, L.~El~Ghaoui, and M.~I. Jordan,
  ``Theoretically principled trade-off between robustness and accuracy,'' in
  \emph{ICML}, 2019.

\bibitem{Tang2019}
C.~Tang, Y.~Fan, and A.~Yezzi, ``An adaptive view of adversarial robustness
  from test-time smoothing defense,'' \emph{arXiv preprint arXiv:1911.11881},
  2019.

\bibitem{Yan2018}
Z.~Yan, Y.~Guo, and C.~Zhang, ``Deep defense: Training dnns with improved
  adversarial robustness,'' in \emph{Advances in Neural Information Processing
  Systems}, 2018, pp. 419--428.

\bibitem{Ros2018}
A.~S. Ross and F.~Doshi-Velez, ``Improving the adversarial robustness and
  interpretability of deep neural networks by regularizing their input
  gradients,'' in \emph{Thirty-second AAAI conference on artificial
  intelligence}, 2018.

\bibitem{Cisse2017}
M.~Cisse, P.~Bojanowski, E.~Grave, Y.~Dauphin, and N.~Usunier, ``Parseval
  networks: improving robustness to adversarial examples,'' in
  \emph{Proceedings of the 34th International Conference on Machine
  Learning-Volume 70}, 2017, pp. 854--863.

\bibitem{Sokolic2017}
J.~Sokoli{\'c}, R.~Giryes, G.~Sapiro, and M.~R. Rodrigues, ``Robust large
  margin deep neural networks,'' \emph{IEEE Transactions on Signal Processing},
  vol.~65, no.~16, pp. 4265--4280, 2017.

\bibitem{Gu2015}
S.~Gu and L.~Rigazio, ``Towards deep neural network architectures robust to
  adversarial examples,'' 2015.

\bibitem{das2018shield}
N.~Das, M.~Shanbhogue, S.-T. Chen, F.~Hohman, S.~Li, L.~Chen, M.~E. Kounavis,
  and D.~H. Chau, ``Shield: Fast, practical defense and vaccination for deep
  learning using jpeg compression,'' in \emph{Proceedings of the 24th ACM
  SIGKDD International Conference on Knowledge Discovery \& Data Mining}, 2018,
  pp. 196--204.

\bibitem{guo2018countering}
C.~Guo, M.~Rana, M.~Cisse, and L.~van~der Maaten, ``Countering adversarial
  images using input transformations,'' in \emph{International Conference on
  Learning Representations}, 2018.

\bibitem{raff2019barrage}
E.~Raff, J.~Sylvester, S.~Forsyth, and M.~McLean, ``Barrage of random
  transforms for adversarially robust defense,'' in \emph{Proceedings of the
  IEEE Conference on Computer Vision and Pattern Recognition}, 2019, pp.
  6528--6537.

\bibitem{shafahi2019adversarially}
A.~Shafahi, P.~Saadatpanah, C.~Zhu, A.~Ghiasi, C.~Studer, D.~Jacobs, and
  T.~Goldstein, ``Adversarially robust transfer learning,'' in
  \emph{International Conference on Learning Representations}, 2019.

\bibitem{zagoruyko2016paying}
S.~Zagoruyko and N.~Komodakis, ``Paying more attention to attention: Improving
  the performance of convolutional neural networks via attention transfer,''
  \emph{arXiv preprint arXiv:1612.03928}, 2016.

\bibitem{zhang2019towards}
H.~Zhang, H.~Chen, C.~Xiao, S.~Gowal, R.~Stanforth, B.~Li, D.~Boning, and C.-J.
  Hsieh, ``Towards stable and efficient training of verifiably robust neural
  networks,'' in \emph{International Conference on Learning Representations},
  2019.

\bibitem{webb2018statistical}
S.~Webb, T.~Rainforth, Y.~W. Teh, and M.~P. Kumar, ``A statistical approach to
  assessing neural network robustness,'' in \emph{International Conference on
  Learning Representations}, 2018.

\bibitem{lecuyer2019certified}
M.~Lecuyer, V.~Atlidakis, R.~Geambasu, D.~Hsu, and S.~Jana, ``Certified
  robustness to adversarial examples with differential privacy,'' in \emph{2019
  IEEE Symposium on Security and Privacy (SP)}.\hskip 1em plus 0.5em minus
  0.4em\relax IEEE, 2019, pp. 656--672.

\bibitem{cohen2019certified}
J.~Cohen, E.~Rosenfeld, and Z.~Kolter, ``Certified adversarial robustness via
  randomized smoothing,'' in \emph{International Conference on Machine
  Learning}, 2019, pp. 1310--1320.

\end{thebibliography}

\begin{IEEEbiography}
[{\includegraphics[width=1in,height=1.3in,clip,keepaspectratio]{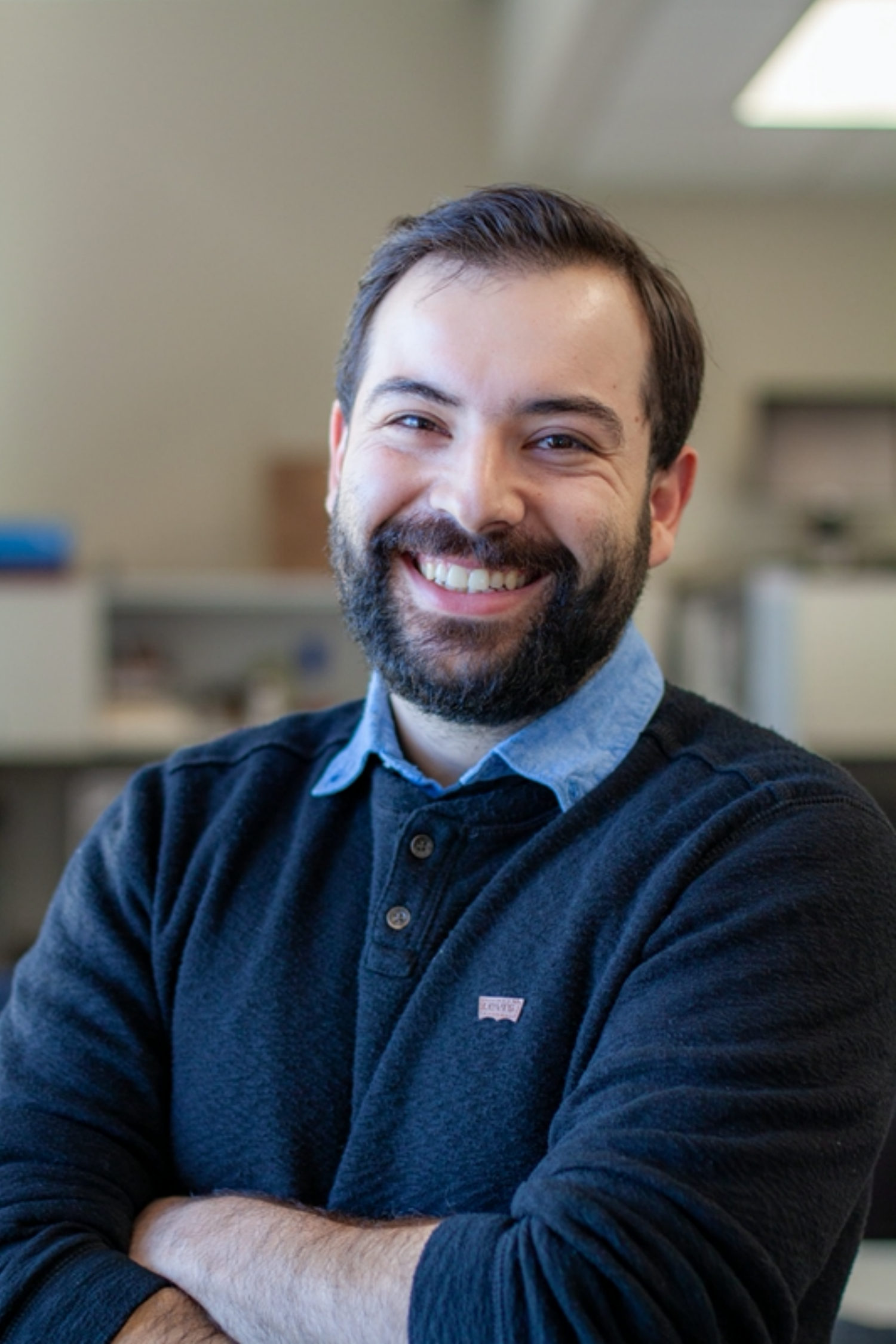}}]{Samuel Henrique Silva}
is currently a Ph.D. student and research fellow at the Open Cloud Institute of University of Texas at San Antonio (UTSA), San Antonio, TX, USA. Samuel received the Bachelor of Science (B.Sc.) degree in Control and Automation Engineering from State University of Campinas, Campinas, Brazil, in 2012 and the M.S. degree in Electrical Engineering from the University of Notre Dame, Notre Dame, IN, USA in 2016. He is a member of the IEEE, Eta Kappa Nu honor society. Samuel's research interests are in the areas of artificial intelligence, robustness in deep learning models, autonomous decision making, multi-agent systems and adversarial environments.
\end{IEEEbiography}

\begin{IEEEbiography}
[{\includegraphics[width=1in,height=1.3in,clip,keepaspectratio]{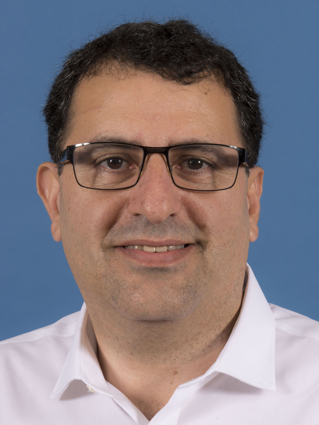}}]{Paul Rad}
is a co-founder and Associate Director of the Open Cloud Institute (OCI), and an Associate Professor with the Information Systems and Cyber Security Department at the University of Texas at San Antonio. He received his first B.S. degree from Sharif University of Technology in Computer Engineering in 1994, his 1st master in artificial intelligence from the Tehran Polytechnic, his 2nd master in computer science from the University of Texas at San Antonio (Magna Cum Laude) in 1999, and his Ph.D. in electrical and computer engineering from the University of Texas at San Antonio. 
He was a recipient of the Most Outstanding Graduate Student in the College of Engineering, 2016, earned the Rackspace Innovation Mentor Program Award for establishing Rackspace patent community board structure and mentoring employees (2012), earned the Dell Corporation Company Excellence (ACE) Award  for exceptional performance and innovative product research and development contributions (2007), and earned the Dell Inventor Milestone Award, Top 3 Dell Inventor of the year (2005). He holds 15 U.S. patents on cyber infrastructure, cloud computing, and big data analytics with over 300 product citations by top fortune 500 leading technology companies such as Amazon, Microsoft, IBM, Cisco, Amazon Technologies, HP, and VMware. He has advised over 200 companies on cloud computing and data analytics with over 50 keynote presentations. High performance cloud group chair at the Cloud Advisory Council (CAC), OpenStack Foundation Member (the \#1 open source cloud software), San Antonio Tech Bloc Founding Member, and Children’s Hospital of San Antonio Foundation board member.
\end{IEEEbiography}

\end{document}